# Smart Buildings Energy Consumption Forecasting using Adaptive Evolutionary Ensemble Learning Models

Mehdi Neshat[1*], Menasha Thilakaratne[2], Mohammed El-Abd[3], Seyedali Mirjalili[4,5], Amir H. Gandomi[1,5], and John Boland[6]

[1] *Faculty of Engineering and Information Technology, University of Technology Sydney, Sydney, 2007, NSW, Australia*
mehdi.neshat@uts.edu.au; gandomi@uts.edu.au

[2] *School of Computer Science, The University of Adelaide, Adelaide, 5005, Australia* menasha.thilakaratne@adelaide.edu.au

[3]*College of Engineering and Applied Sciences, American University of Kuwait, Kuwait, melabd@auk.edu.kw*

[4]*Center for Artificial Intelligence Research and Optimization, Torrens University Australia, Brisbane, QLD 4006, Australia,ali.mirjalili@laureate.edu.au*

[5]*University Research and Innovation Center (EKIK), Obuda University, Budapest, 1034, Hungary*

[6]*Industrial AI Research Centre, UniSA STEM, University of South Australia, Mawson Lakes, 5095, Australia*
*John.Boland@unisa.edu.au*

**Abstract**

Smart buildings are gaining popularity because they have the capability to enhance the energy efficiency of buildings, lower costs, improve security, and provide a more comfortable and convenient environment for the people occupying the building. A considerable ratio of the global energy supply has been consumed in building sectors and plays a pivotal role in the future decarbonisation pathways. In order to manage energy consumption and improve energy efficiency systems in smart buildings, developing reliable and accurate energy demand forecasting is crucial and meaningful. However, extending an effective predictive model for the total energy use of appliances at the buildings' level is challenging because of temporal oscillations and complex linear and non-linear patterns. This paper proposes three hybrid ensemble predictive models, incorporating Bagging, Stacking, and Voting mechanisms combined with a fast and effective evolutionary hyper-parameters tuner. The performance of the proposed energy forecasting model was evaluated using a hybrid dataset of meteorological parameters, energy use of appliances, temperature, humidity, and lighting energy consumption of different sections collected by 18 sensors in a building which is located in Stambruges, Mons in Belgium. In order to provide a comparative framework and investigate the efficiency of the proposed predictive model, 15 popular machine learning (ML) models, including two classic ML models, three Neural Networks (NN), a Decision Tree (DT), a Random Forest (RF), two Deep Learning (DL) and six Ensemble models, were compared. The prediction results indicate that the adaptive evolutionary bagging model surpassed other predictive models in both accuracy and learning error. Notably, it delivered accuracy gains of 12.6%, 13.7%, 12.9%, 27.04%, and 17.4% when compared to Extreme Gradient Boosting (XGB), Categorical Boosting (CatBoost), Gradient Boosting Machine (GBM), Light Gradient Boosting Machine (LGBM), and RF.

**Nomenclature**

Table 1: Summary of key abbreviations used in the manuscript for clarity.

| abbreviation | fullname |
|---|---|
| AI | Artificial intelligence |
| ANN | Artificial Neural networks |
| Bi-LSTM | Bidirectional Long short-term memory network |
| BIM-DB | building information modeling-design builder |
| BIM | Building Information Modeling |
| BS | Batch size |
| CART | Classification and regression tree |
| CatBoost | Categorical Boosting |
| CR | Probability crossover rate |
| CL | Cooling load |
| CMA-ES | Covariance matrix adaptation evolution strategy |
| CNN | Convolutional neural network |
| DDPG | Deep Deterministic Policy Gradient |
| DE | Differential evolution |
| DNN | Deep neural networks |
| DT | Decision Tree |
| EA | Evolutionary Algorithm |
| ELM | Extreme Learning Machine |
| EVS | Explained variance score |
| GA | Genetic algorithm |
| GBT | Gradient boosting tree |
| GBM | Gradient Boosting Machine |
| GC | Generalised correntropy |
| GPT | Generative Pre-trained Transformers |
| GRU | Gated recurrent unit |
| HGBR | Histogram-Based Gradient Boosting Regressor |
| HL | Heating load |
| HVAC | Heating, Ventilation, and Air Conditioning |
| IoT | Internet of Things |
| LGBM | Light Gradient Boosting Machine |
| LOF | Local outlier factor algorithm |
| LRD | Local reachability density |
| LHTES | Latent heat thermal energy storage |
| LSTM | Long short-term memory network |
| MAE | Mean absolute error |
| ML | Machine learning |
| MLP | Multi-layer perceptron |
| MSE | Mean square error |
| NSGA | Non-dominated Sorting Genetic Algorithm |
| NM | Nelder-Mead simplex direct search method |
| PSO | Particle Swarm Optimisation |
| PHPP | Passive House Planning Package |
| RF | Random Forest |
| RIME | Rime optimisation algorithm |
| RMSE | Root mean square error |
| RNN | Recurrent neural networks |
| SCO | Sine cosine optimisation |
| SMAPE | Symmetric mean absolute percentage error |
| SVM | Support vector machines |
| XGB | Extreme Gradient Boosting |

# 1. Introduction

One-third of the world's primary energy is approximately consumed by buildings [1]. Buildings are a significant contributor to carbon dioxide ($CO_2$) emissions, accounting for nearly 39% of such emissions [2]. Due to this high level of buildings' energy consumption contribution to global energy demand, developing smart buildings is crucial. There are numerous advantages in advancing smart buildings, such as enhanced energy optimisation, augmented residents' satisfaction and productivity [3], as well as improved health and well-being [4]. These benefits have been achieved due to hiring cutting-edge technologies such as artificial intelligence (AI)-based methods, deep neural networks (DNNs) [5], and adaptive learning controls in smart buildings [6], which enable such facilities to control various systems (cooling, heating, cooking, etc. [7]) to evolve more efficient in terms of energy and comfort [8]. Furthermore, smart buildings prioritise indoor air quality, ensuring thermal, acoustic, and visual comfort. In smart buildings, to enhance communication and information sharing, incorporating technologies have been used, like the Internet of Things (IoT), Building Information Modeling (BIM), and Blockchain conducted to improve security and management [9]. Another significant advantage of developing smart buildings is contributing to the energy sector decarbonisation [10] by supporting the electrical grid through providing demand response functionality [11] and balancing electricity demand with non-dispatchable renewable energy sources [12].

In the last two decades, various ML techniques have experienced significant growth, particularly in modelling energy consumption in smart buildings. This surge of interest can be attributed to the remarkable efficacy and robustness exhibited by ML predictors in this field. Impressively, ML models have demonstrated exceptional generalisation and flexibility abilities [13], making them widely pertinent to a diverse range of problems. They have been hailed as "universal function approximators" because of their unparalleled adaptability. A comprehensive review of the rapid advancements in Artificial Intelligence (AI) and ML models within the context of smart buildings has yielded a meaningful conclusion [14] and determined that the overall adaptability of buildings to unforeseen changes can be significantly enhanced through the enactment of AI-driven learning processes. Moreover, integrating adaptability solutions at the timescales of heating, ventilation, and air conditioning (HVAC) control and electricity market participation has been identified as the most promising avenue for achieving substantial improvements in energy efficiency.

One pivotal advantage of employing ML models lies in their aptitude for analysing extensive datasets and uncovering intricate patterns that elude traditional statistical methodologies. By considering an array of factors, such as construction characteristics, occupancy patterns, and weather states, these models offer accurate predictions of energy usage within buildings [15]. This capability stems from their capacity to process vast volumes of data and discern hidden correlations that may remain inconspicuous otherwise. Moreover, the prevalence of multiple sensors for data collection in smart buildings necessitates the development of real-time systems for monitoring, controlling, predicting, and optimising total power consumption. ML models excel in this arena by continuously analysing sequential data and constructing precise models of these dynamic systems [16]. Through incessant monitoring and data analysis, these models can adapt control settings for Heating, Ventilation, and Air Conditioning (HVAC) systems, lighting, and other building components to attain desired energy efficiency targets. Recently, Lie et al. [17] proposed a novel HVAC control system for intelligent buildings that uses a multi-step predictive deep learning model to reduce power consumption costs while maintaining user satisfaction. The system combines Long Short-term Memory (LSTM), generalised correntropy (GC) loss function, and Deep Deterministic Policy Gradient (DDPG) for predicting house temperature and dynamic power adjustment. Simulation results showed over 12% cost savings compared to alternative approaches.

Another compelling rationale for incorporating ML models in energy demand modelling for smart buildings lies in their forecasting capabilities; by leveraging historical data, weather forecasts, and other pertinent characteristics, ML aids in predicting future energy demands accurately [18]. This proficiency in demand forecasting facilitates superior planning for energy generation, distribution, and load management, culminating in a more dependable and efficient energy supply. These factors collectively enable the optimisation of energy utilisation, enhance operational efficiency, and contribute to the establishment of sustainable [19] and intelligent building systems.

Somu et al. [20] proposed a hybrid building power consumption model (kCNN-LSTM) consisting of LSTM, a Convolutional neural network (CNN) combined with a K-means clustering method and sine cosine optimisation (SCO) algorithm [21] to tune the hyper-parameters of LSTM. The kCNN-LSTM model outperforms existing demand forecast models and offers precise energy consumption. An automated building energy load forecasting methodology [22] has recently been introduced based on Generative Pre-trained Transformers (GPT) in combination with prompt optimisation, external knowledge use, and self-correction. The method severely mitigates technical barriers to entry for non-experts and permits precise low-budget energy prediction. It was compared with actual test buildings and proved to have a mean R2 of 0.95, demonstrating the engineering viability of mass language models for smart building energy management innovation.

While ML models have been shown to be promising for the prediction of building energy consumption, current research focuses mainly on short-term prediction and seldom introduces new parameters to improve the accuracy of predictions. To address this gap, a team developed a data-driven method [23] to predict the hourly energy consumption of a university office building by integrating meteorological, temporal, and an introduced meta-parameter—air conditioning demand. Five ML algorithms (Random Forest (RF), Gradient Boosted Trees (GBT), Support Vector Machines (SVM), Artificial Neural Networks (ANN), and Deep Neural Networks (DNN)) are compared and experimental results show that DNN provide the best performance (Root mean square error (RMSE) = 4.796 kWh, Mean Absolute Percentage Error (MAPE) = 5.738%), outperforming existing methods. Incorporating the air conditioning demand parameter significantly enhances model accuracy for every algorithm.

Ensemble models offer excellent benefits in building energy prediction [24] by exploiting the strengths of different algorithms, enhancing prediction accuracy and generalisability compared to individual models. While much attention is being given now, most prior studies have focused on single ML models or basic ensemble techniques without fully harnessing stacked architectures for heating and cooling load (HL and CL) prediction. Furthermore, less research has been done in the literature on integrating hyperparameter tuned models with heterogeneous base models for residential building energy prediction. Closing these gaps, in a recent work [25], a stacked ensemble model was introduced integrating XGB, DT, RF, and Bayesian optimisation for hyperparameter tuning. The suggested model performed considerably better than the traditional techniques, providing better performance (RMSE of 0.484 for HL and 0.948 for CL). Another example of ensemble models is [26] proposing a stacked learning model for predicting the dynamic performance of PCM-based double-pipe latent heat thermal energy storage (LHTES) units. Main contributions include sensitivity analysis for variable selection, a two-stage ensemble model combining Regression Trees, SVR, and Linear Regression, and comprehensive validation over datasets and phase change stages. The infrastructure would be able to enhance 7.82% more MAPE, make 25.6% greater stability, and achieve 9.7% peak reduction demand in heating, ventilation, and air conditioning (HVAC) systems, moving toward flexible data-driven building energy management. Another study [27] suggested a stacking ensemble learning model for home net load-interval prediction, which combines k-means user clustering, LRIME-based optimisation, and bootstrap interval estimation. Their main contributions included developing interpretable interval forecasts, recommending the rime optimisation algorithm (LRIME) for improved performance, and adding LSTM, XGBoost, and ELM as optimised base learners. Australian Ausgrid data tests confirm the model's improved accuracy, robustness, and uncertainty estimation over state-of-the-art models.

Combining ML models with optimisation methods is one of the popular techniques used to forecast energy consumption in buildings. To address the lack of integrated prediction and optimisation methods in green building design, a recent study [28] proposed a framework combining BIM-DB simulation, Bayesian-Random Forest (Bayesian-RF) prediction, and Non-dominated Sorting Genetic Algorithm (NSGA-III) optimisation. BIM-DB efficiently generates building performance data, while Bayesian-RF achieves high prediction accuracy (*MSE*

< 0.08, $R2 > 0.85$). The prediction model guides NSGA-III to optimise energy use, emissions, cost, and thermal comfort. A teaching building case study shows reductions of 7.68% in energy consumption, 6.48% in carbon emissions, and 1.77% in cost, along with improved comfort. Current approaches to optimising public building sustainability tend to find it difficult to reconcile competing goals and combine expert knowledge with data-driven forecasting. A recent study [29] suggested a hybrid approach that blended building information modeling-design builder (BIM-DB) simulations with a BO-CatBoost-NSGA-III algorithm to overcome these limitations. Their major contributions included a two-stage knowledge–data-driven approach to secure dataset generation, a BO-optimised CatBoost model with $R2 > 0.97$ across targets, and finally, multiobjective optimisation using NSGA-III, which delivered 32.20% lower energy consumption,
48.77% lower $CO_2$ emissions, 60.69% improved thermal comfort, and 15.45% less glare.

Sequential ML models, such as LSTM, BiLSTM, CNN-LSTM, etc., have gained recognition for their success in these specific domains [30]. However, they do come with certain drawbacks that need to be considered as follows.

- One notable disadvantage is the complex architecture of these models, which can result in extensive training runtimes, mainly when dealing with large-scale datasets. Consequently, the computational requirements for training these models can be substantial.

- Moreover, achieving optimal performance with these models heavily relies on careful design and parameter tuning. Improper choice of hyper-parameters can lead to suboptimal performance or overfitting, underscoring the need for meticulous attention during the model configuration phase.

- Another drawback is the need for more interpretability of LSTM and its family models. These models are often considered black boxes, making comprehending the underlying reasoning behind their predictions challenging. Interpreting the learned representations and understanding the critical features becomes a non-trivial task.

- Furthermore, when faced with limited data, these sequential models may struggle to extract meaningful patterns and achieve optimal performance [31]. Uncovering hidden patterns and dependencies relies heavily on the availability of sufficient training examples, which can be a limitation in scenarios where data is scarce.

Considering these drawbacks is crucial when deciding whether to employ LSTM, BiLSTM, or CNNLSTM models. The trade-off between their success in specific domains and the associated challenges of training runtime, parameter tuning, interpretability, and data limitations should be carefully evaluated to ensure the most suitable approach for a given application.

To address the aforementioned challenges, in this study, we propose a hybrid learning model specifically designed for predicting the total power usage of compliances in a Stambruges, Mons, Belgium building. The model incorporated three ensemble mechanisms: Bagging, Stacking and Voting models and a fast and effective Evolutionary framework. The study's primary objective was to develop a robust and accurate power consumption prediction model for smart buildings. To achieve this, data collected from 18 sensors installed in the building was used to capture meteorological parameters, energy use of appliances, temperature, humidity, and lighting energy consumption of different sections. The main contributions of this study are listed as follows:

- Comprehensive data analysis was conducted to extract various characteristics and correlations among the collected features and power consumption. This analysis provided valuable insights into the relationships between different variables, helping to inform the development of the predictive model. • A wide range of machine and deep learning models were implemented and compared to ensure the most efficient learning model. This included classic ML models such as DT and RF, as well as various Neural Networks (NN) and Ensemble models. By developing this comprehensive comparative framework, the designers will be able to identify the most effective learning model for predicting power consumption in the smart building context.

- Further, the study addressed the challenge of hyper-parameter tuning initialisation, which can significantly impact the model's performance. To overcome this challenge, four optimisation methods were tested and compared to improve prediction accuracy and reduce modelling training errors. The aim was to find a practical and smart hyper-parameter tuner that would enhance the overall performance of the power consumption prediction model.

- Finally, this study contributes to the field of smart buildings by proposing an adaptive evolutionary ensemble learning model that leverages the power of various ML and tree-based techniques combined with a fast and effective Evolutionary algorithm. To this end, we developed and evaluated six Voting models, eight Bagging models, and ten Stacking architectures, each composed of different configurations of decision trees, gradient-boosted methods, and neural learners. The comprehensive data analysis, extensive model comparison, and optimisation methods employed in this study provide valuable insights and techniques for accurately predicting power consumption in similar smart building scenarios.

In this study, we commence by introducing the dataset utilised, followed by a statistical analysis aimed at unveiling concealed data characteristics (Section 2). Subsequently, we expound upon the technical aspects of the employed methods, encompassing optimisation techniques, the XGBoost model, and adaptive evolutionary ensemble algorithms (Section 3). Subsequent to this, we present the numerical results and engage in a comprehensive discussion (Section 4) to discern the efficacy and efficacy of our proposed method. Ultimately, we summarise our findings, emphasising the advantages of our approach (Section 6).

## 2. Data sets and statistical analysis

The hybrid dataset utilised in this study was obtained from a residential property in Stambruges, Belgium, approximately 24 $km$ from the City of Mons [32]. The house's construction was completed in December 2015, incorporating entirely new mechanical systems. The architectural design followed the principles of passive house certification [33], which entails limiting the annual heating and cooling loads to a maximum of 15 $kWh/m^2$ per year, as determined by design software (Passive House Planning Package (PHPP)). It is worth highlighting that in September 2016, the building's air leakage was assessed and measured to be 0.6 air changes per hour at 50 $Pa$. A heat recovery ventilation unit with an efficiency ranging between 90% and 95% is employed to ensure proper ventilation. The total floor area of the house amounts to 280 $m^2$, with the heated area encompassing 220 $m^2$. The map of two floors of the building [32] with the location of sensors to record temperature and humidity.

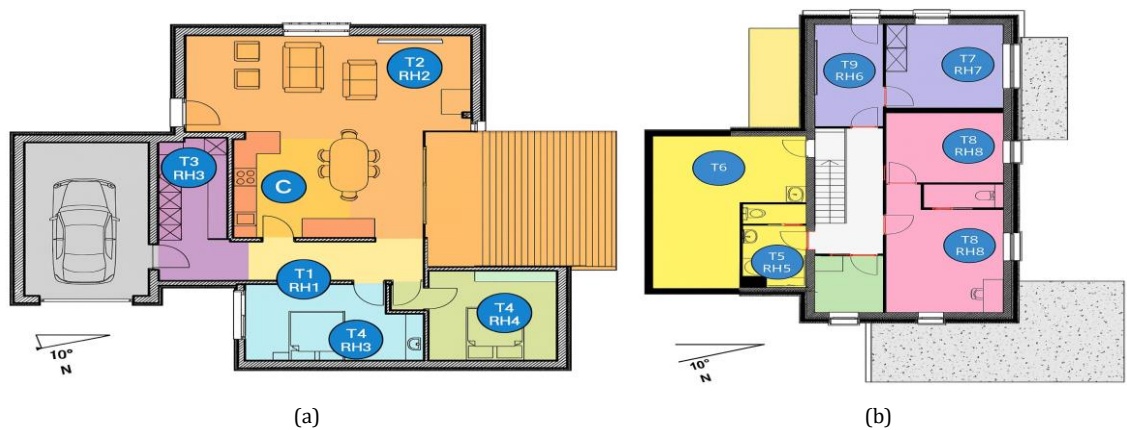

(a)          (b)

Figure 1: The building map of (a) the First and (b) the Second floor and temperature and humidity sensors position.

Electrical energy consumption in the passive house was monitored using M-BUS energy counters, capturing data every 10 minutes. This tracking included individual power loads from the domestic hot water, devices, lighting, heat recovery ventilation unit, and electric baseboard heaters. The energy devices used correspond to the list given in reference [32]. An internet-based energy monitoring system collects the energy data, keeps it and dispatches notifications via email every 12 hours. Lighting energy consumption constituted between 1% and 4% of the total, predominantly due to LED fixtures. Temperature and humidity conditions within the house were tracked using a wireless sensor network (ZigBee) constructed with XBeeradios, Atmega328P microcontrollers, and DHT-22 sensors. The house's large size and solid construction necessitated the inclusion of two additional XBee radios functioning as routers to facilitate effective communication from the end nodes to the coordinator. Battery-powered sensor nodes relayed information approximately every 3.3 minutes. The list of variables with their location in the dataset can be seen in Table S1.

Table 2 shows a statistical analysis of the dataset's variables and briefly outlines the dataset, signifying key characteristics such as coverage, prominent tendency, and variability.

Table 2: Statistical analysis of total energy consumption of the building and other features.

|  | Appliances | lights | T1 | RH 1 | T2 | RH 2 | T3 | RH 3 | T4 | RH 4 | T5 | RH 5 | T6 | RH 6 |
|---|---|---|---|---|---|---|---|---|---|---|---|---|---|---|
| Min | 10.000 | 0.000 | 16.790 | 27.023 | 16.100 | 20.463 | 17.200 | 28.767 | 15.100 | 27.660 | 15.330 | 29.815 | -6.065 | 1.000 |
| Max | 1080.000 | 70.000 | 26.260 | 63.360 | 29.857 | 56.027 | 29.236 | 50.163 | 26.200 | 51.090 | 25.795 | 96.322 | 28.290 | 99.900 |
| Mean | 97.695 | 3.802 | 21.687 | 40.260 | 20.341 | 40.420 | 22.268 | 39.243 | 20.855 | 39.027 | 19.592 | 50.949 | 7.911 | 54.609 |
| Median | 60.000 | 0.000 | 21.600 | 39.657 | 20.000 | 40.500 | 22.100 | 38.530 | 20.667 | 38.400 | 19.390 | 49.090 | 7.300 | 55.290 |
| STD | 102.525 | 7.936 | 1.606 | 3.979 | 2.193 | 4.070 | 2.006 | 3.255 | 2.043 | 4.341 | 1.845 | 9.022 | 6.090 | 31.150 |
|  | T7 | RH 7 | T8 | RH 8 | T9 | RH 9 | T out | Press mm hg | RH out | Windspeed | Visibility | Tdewpoint | rv1 | rv2 |
| Min | 15.390 | 23.200 | 16.307 | 29.600 | 14.890 | 29.167 | -5.000 | 729.300 | 24.000 | 0.000 | 1.000 | -6.600 | 0.005 | 0.005 |
| Max | 26.000 | 51.400 | 27.230 | 58.780 | 24.500 | 53.327 | 26.100 | 772.300 | 100.000 | 14.000 | 66.000 | 15.500 | 49.997 | 49.997 |
| Mean | 20.267 | 35.388 | 22.029 | 42.936 | 19.486 | 41.552 | 7.412 | 755.523 | 79.750 | 4.040 | 38.331 | 3.761 | 24.988 | 24.988 |
| Median | 20.033 | 34.863 | 22.100 | 42.375 | 19.390 | 40.900 | 6.917 | 756.100 | 83.667 | 3.667 | 40.000 | 3.433 | 24.898 | 24.898 |
| STD | 2.110 | 5.114 | 1.956 | 5.224 | 2.015 | 4.151 | 5.317 | 7.399 | 14.901 | 2.451 | 11.795 | 4.195 | 14.497 | 14.497 |

Figure 2 illustrates the distribution of the energy consumption profile over five months. The graph displays a significant variance in energy usage, ranging from zero to 1000 *Wh*. From a broad perspective, no discernible pattern is observed, presenting a challenging scenario for the accurate estimation of power utilisation by ML models.

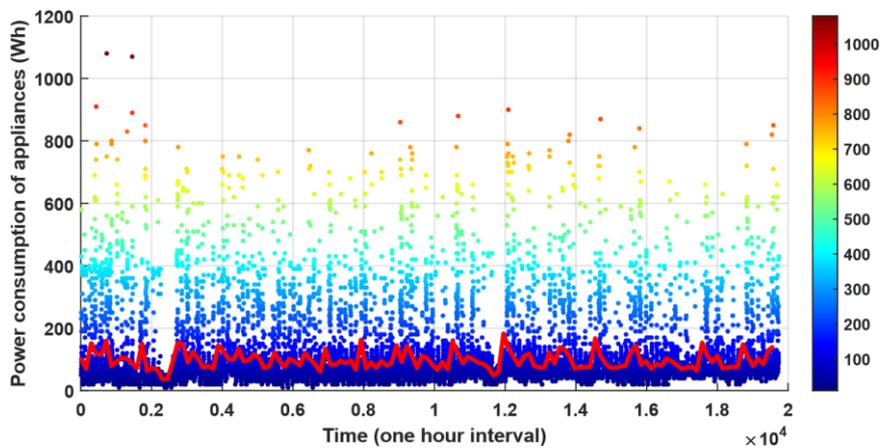

Figure 2: Appliances power consumption observations all-in-one.

Figure 3 is a plot of the daily average time series profiles of temperature and humidity data recorded by nine sensors mounted across the interior and exterior of the smart building. Out of them, T6 and T-out are the outdoor conditions, and the remaining represent indoor climate measurements. The result shows that the indoor sensors display a consistent and stable thermal trend over the four-month observation period, indicating a well-managed indoor environment. On the other hand, T6 and T-out are more diverse, reflecting the effect of outside weather volatility. Overall, the average outdoor temperature, at approximately 15 °C, is considerably lower than indoor temperatures, a reflection of the quality of the building's insulation and the effectiveness of internal climate control.

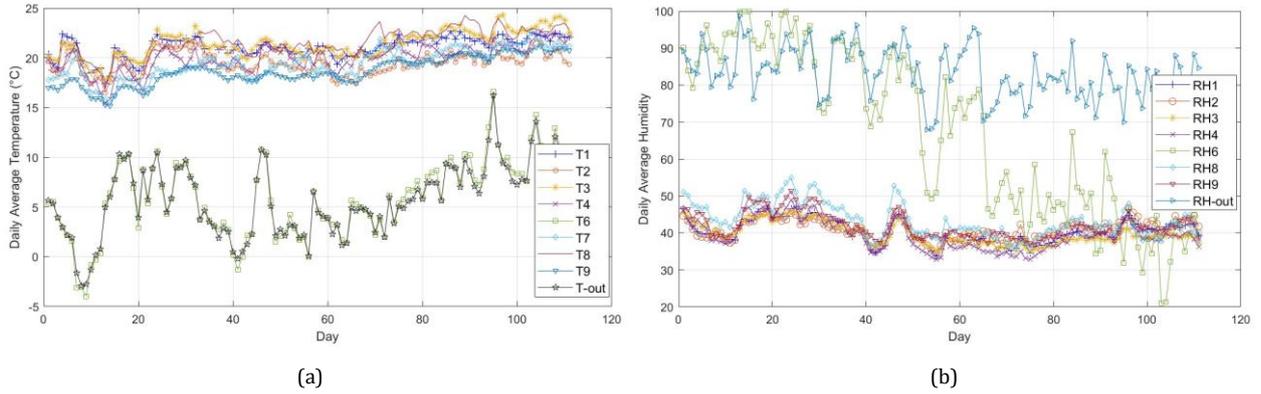

(a)            (b)

Figure 3: Time series of daily average (a) temperature and (b) humidity recorded from sensors.

In Figure 4b, we observe the descriptive statistics of power consumption across the five-month period, specifically from January to May. Remarkably, the average power consumption in January and April is the highest among the months considered. This information provides insights into the varying power usage levels throughout the months. Besides, when comparing weekdays and weekends, Figure 4c reveals that Thursday and Saturday are the days with the highest energy consumption. This data further highlights the distinction between energy consumption patterns on different days of the week. These graphical representations contribute to a comprehensive understanding of the energy consumption dynamics, highlighting the challenges faced by the ML model in accurately estimating power utilisation.

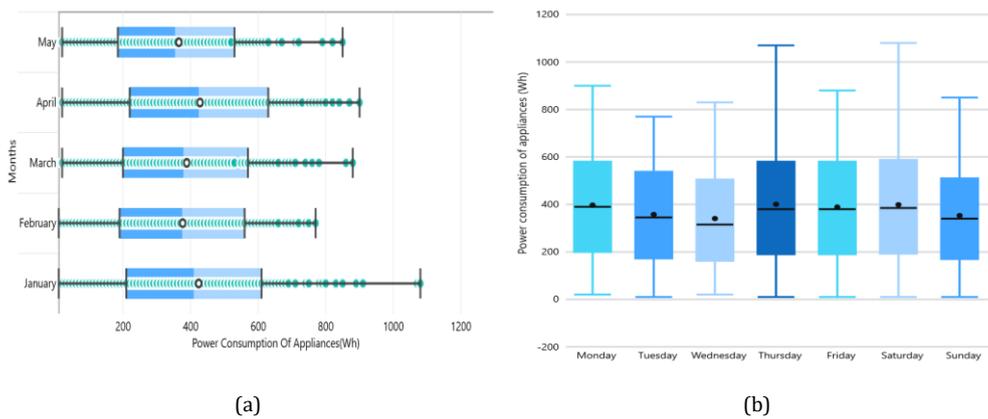

(a)            (b)

Figure 4: (a) The distribution of consumption through five months. (b) The statistical observations for energy consumption in five months as a box-plot.

Figure 5 depicts the average electricity usage of both devices and lights at different times. The graph reveals a considerable correlation between the two variables. Particularly, a high correlation is observed throughout the time range. However, it is noteworthy that between 12:00 PM and 6:00 PM, the average power consumption of devices surpasses that of lights. This finding aligns with expectations, as daytime usage typically involves increased activity and higher demand for device-related electricity. After 6:00 pm, a shift in the pattern becomes evident that the average power consumption of lights increases, likely corresponding to the evening hours when lighting requirements typically become more prominent. Consequently, during this period, the average power consumption of lights surpasses that of devices.

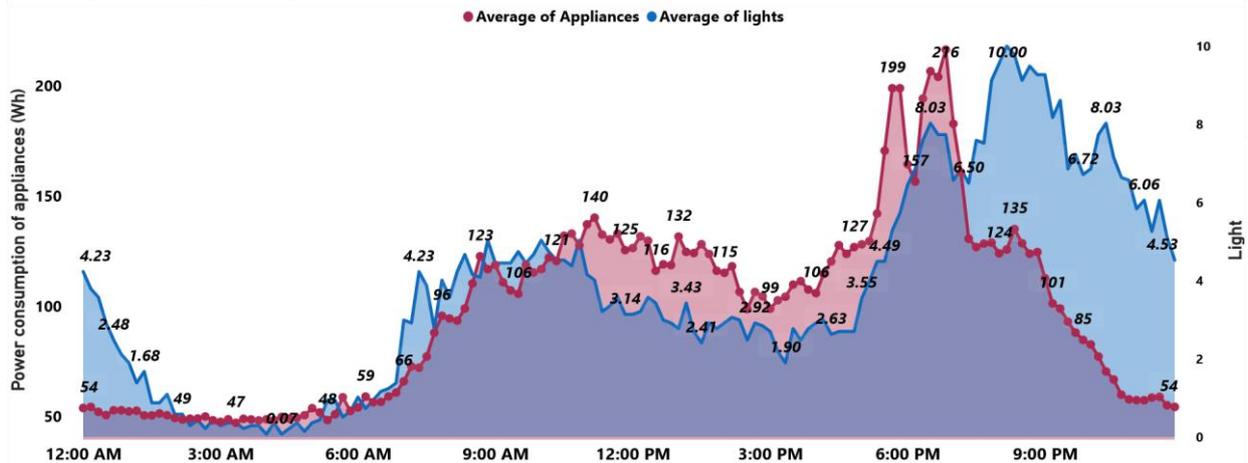

Figure 5: The average power usage of appliances and lights between 12:00 AM and 11:59 PM.

Figure 6 presents the correlation coefficient analysis between temperature variables recorded by ten sensors and the power consumption of appliances. Two noteworthy observations can be made from this analysis. Firstly, a positive correlation is observed between all indoor temperature variables and power consumption. This indicates that as indoor temperatures rise, the power consumption of appliances also tends to increase. Furthermore, there is a positive correlation among the indoor temperature variables themselves, suggesting that similar changes in the others accompany changes in one temperature variable. In contrast, the outdoor temperature variable negatively correlates with power consumption and the other indoor temperature variables. This observation implies that as the outdoor temperature rises, there is an inclination to decline in power consumption and indoor temperatures. This negative correlation likely stems from cooling systems or strategies to maintain comfortable indoor conditions despite higher outdoor temperatures. Last but not least, the highest correlation between appliances and temperature variable T2 indicates a strong relationship between these two factors. Further, the second-largest correlation between appliances and temperature variable T6 is observed, further highlighting their interdependence.

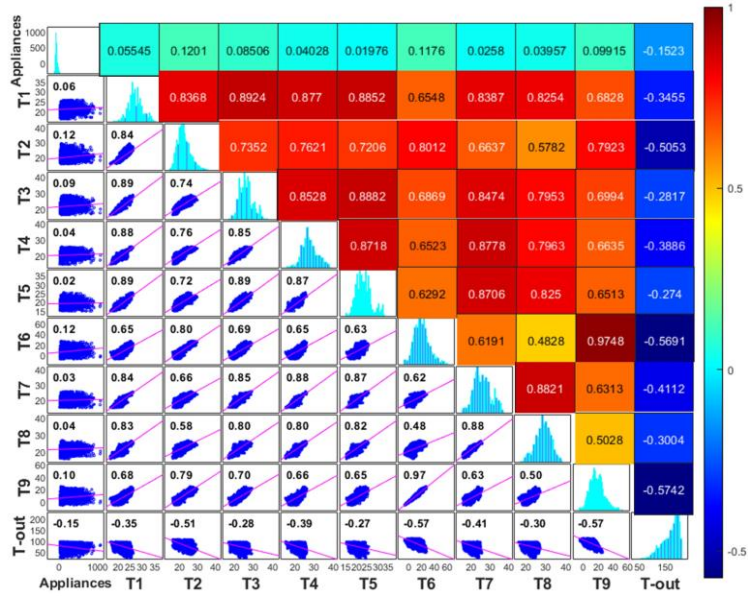

Figure 6: Correlations between the appliances' power consumption and the temperature recorded from nine sensors inside and outside of the building.

To explore the correlation between temperature, humidity variables, and power consumption, we analysed as depicted in Figure 7. This line chart provides insights into the relationships between these variables. The chart reveals a positive correlation pattern among temperature variables, with correlations higher than those observed for humidity features. Nevertheless, most humidity variables exhibit a negative correlation with power consumption, which implies that as humidity levels increase, power consumption tends to decrease. The negative correlations observed for humidity variables highlight the influence of humidity on energy usage patterns. This negative correlation could be attributed to the impact of moisture on cooling requirements, ventilation systems, or other factors affecting power consumption.

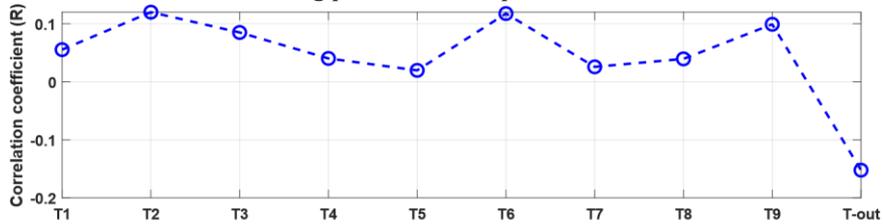

(a)

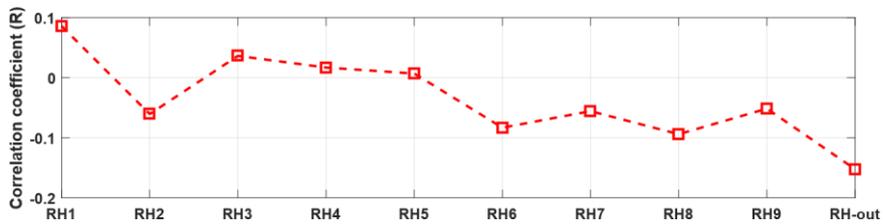

(b)

Figure 7: The correlation between the total power consumption of appliances and (a) nine temperature variables, and (b) humidity variables recorded by wireless sensors.

## 3. Methods

In this section, the technical approaches adopted in this research are presented. Firstly, the Local Outlier Factor algorithm is presented (Section 3.1) to filter and remove outlying data points and present a highquality dataset for model building. Secondly, the meta-heuristic algorithms (Section 3.2), including GA, DE, and the (1+1) Evolutionary Algorithm, and their details in optimisation and search abilities are emphasised. Next, ensemble learning strategies (Section 3.3) such as Stacking, Bagging, Voting, and Boosting are outlined and share their advantages in predictive precision, stability, and generalisability. Finally, this study's novelty, the Adaptive Evolutionary Ensemble Learning model (Section 3.4), is introduced, describing its advantages over ensemble learning and evolutionary algorithms for minimising the function under adverse optimisation landscapes.

*3.1. Local outlier factor (LOF) algorithm*

In order to detect and remove outliers, we used the LOF method [34], which is one of the most popular and effective techniques in time series data cleaning. LOF is an unsupervised, neighbourhood-based algorithm and compares each observation with k-nearest Neighbors estimates, finding the ratio density that estimates the local reachable of observation versus that over its neighbourhood; therefore, it calculates this LOF score, corresponding to an observation's average density to those neighbours. Thues, it considers outlier points whose densities are much lower than their neighbours, which is why LOF effectively finds anomalies within datasets with varying density distributions. Equation 1 shows the LOF computed for $x$ observation [35]. Also, variable $o$ is an observation to an individual nearest observation from among the k-nearest neighbours of data point $x$.

$$LOF_i(x) = \frac{1}{|N_i(x)|} \sum_{o \in N_i(x)}^{N} \frac{\text{LRD}\,dis_i(o)}{LRD_i(x)}, \qquad (1)$$

where the local reachability density shows by *LRD* and $|N_i(x)|$ denotes the number of samples in the neighbourhood of $x$ observation. To compute the rate of reachability distance for each sample in the dataset, Equation 2 was introduced.

$$\widetilde{dis}_i(x,o) = \max(dis_i(o), dis_i(x,o)), \qquad (2)$$

It is noted that $dis_i(o)$ mentions the shortest distance among the neighbours of observation $o$. Therefore, the LRD of observation $x$ is defined as follows.

$$\text{LRD}_i(x) = 1 / \frac{\sum_{o \in N_i(x)} \widetilde{dis}_i(x,o)}{|N_i(x)|} \qquad (3)$$

The formula is calculated as the inverse of the average reachability distance between $x$ and its $k$-nearest neighbours $o \in N_i(x)$. The term $dis_i(x,o)$ represents the reachability distance from $x$ to neighbour $o$, which accounts for both the actual distance and the neighbourhood radius of $o$. This measure quantifies how densely $x$ is located with respect to its local neighbourhood - higher LRD values indicate that $x$ resides in a denser region.

The LOF algorithm is suited for detecting outliers in datasets, including different distributions concerning density, because it uses a relative measure of the density at every point concerning its surrounding neighbours instead of a general threshold value [36]. LOF is also resistant to differing data scales and able to handle both clustered and nonuniform data.

*3.2. Meta-heuristics*

Meta-heuristic algorithms have proved highly effective in optimising the performance of hybrid learning strategies such as ensemble models. These algorithms, namely GA, DE, and (1+1)EA, are optimally employed in challenging optimisation problems where traditional gradient-based or exhaustive search strategies are not applicable [37]. Ensemble learning algorithms typically have several base learners and a number of control parameters, such as learning rates, tree depths, and voting weights, whose manual tuning is time-consuming and inefficient. Meta-heuristic algorithms solve this issue by intelligently searching and exploring the parameter space in order to avoid premature convergence, making a balance between global and local search [38]. Their ability to operate without derivative information and adapt to very nonlinear, high dimensional objective landscapes render them highly beneficial for hyperparameter optimisation by hand and the improvement of ensemble model accuracy, stability, and generalisation.

*3.2.1. Differential Evolution (DE) algorithm*

DE [39] is an evolutionary computational method, population-based, inspired by biological processes that use a stochastic search strategy to find the global optimum of a given problem. DE generates and maintains a population of candidate solutions, and each solution is designated as a vector of decision variables (binary, discrete or continuous values) in the optimisation problems. In order to evaluate the fitness of each solution, an objective function is introduced and based on this fitness, the solutions can be sorted. In the following generations, DE algorithm develops new vectors (offspring) by integrating and mutating individuals in the current population. The primary evolutionary operators of DE include crossover, mutation and selection.

*Mutation Operator:.* In DE algorithms, the most significant operator is the mutation that stochastically perturbs a solution in the population to generate a new candidate solution [40]. The popular type of DE mutation is entitled "DE/rand/1/bin" (Equation 4). This strategy often intuitively supports the stronger exploration ability but almost shows a low convergence speed , promoting global exploration and reducing the risk of premature convergence. As a result, this strategy can usually be used to optimise problems with multi-modal attributes.

$$\vec{V}_g = \vec{D}_{r1} + \omega \times (\vec{D}_{r2} - \vec{D}_{r3}) \tag{4}$$

where $\vec{V}_g$ is the differential vector of three candidates ($\vec{D}_{r1}$, $\vec{D}_{r2}$, and $\vec{D}_{r3}$) chosen randomly from the current population, and to tune the exploration step size, $\omega$ is introduced as the mutation factor.

*Crossover Operator:.* The binomial crossover strategy of DE enjoys several advantages that result in its effectiveness and wide use in real-world problems of continuous optimisation. Its computationally efficient and simple construction relies on random sampling and component-wise replacement, hence making it scalable to high-dimensional problems. Crossover rate ($C_R$) is a direct control parameter that facilitates flexible balancing between exploitation and exploration by regulating the fraction of the mutant vector in the trial solution. This promotes population diversity and avoids premature convergence. In addition, binomial crossover usually includes a mechanism to ensure that at least one component of the mutant vector is incorporated into the trial vector to prevent cyclical solutions and enhance the local optima avoidance ability of the algorithm. Its generality in various problem spaces and robustness to diverse objective function topologies also speak volumes about its effectiveness in solving complicated optimisation problems. The crossover operator combines the mutated solution with another one in the current population to form a trial solution. One of the well-known types of crossover is binomial [40], formulated based on Equation 5.

$$\vec{S}^{i,j} = \begin{cases} \vec{V}^{i,j} & \text{if } (r \leq C_R) \text{ or } (j == C_n), j = 1,2, \dots N_D \\ \vec{D}^{i,j} & \text{otherwise.} \end{cases} \tag{5}$$

where $S$ and $C_R$ are the trial vector and the rate of probability crossover defined in the range of [0-1], respectively. $C_n$ is the index of solutions chosen in the crossover.

*Selection Strategy:.* In Differential Evolution (DE), the selection strategy plays a crucial role in guiding the evolution process to optimal solutions. After a trial vector is created through mutation and crossover, DE applies a greedy selection strategy to determine whether the new solution should be retained. The new solution ($S^i$) is generated and combined with its parent ($D^i$) to replace the offspring as follows.

$$\vec{D}_i^{g+1} = \begin{cases} \vec{S}_i^g & \text{if } f(\vec{S}_i^g) \le f(\vec{D}_i^g) \\ \vec{D}_i^g & \text{otherwise.} \end{cases} \tag{6}$$

DE exhibits outstanding power in solving optimisation problems and has advantages such as simplicity, reliability, and robustness, and is particularly useful for solving complex optimisation problems where the objective function is non-linear, non-convex [41] and may have multiple local optima. However, DE has weaknesses, including slow convergence speed, difficulty adjusting parameters for different problems, and performance deterioration with increasing search space dimensionality.

*3.2.2. Genetic Algorithms (GA)*

GAs are population-based stochastic optimisation techniques that emulate the process of evolutionary biology to identify the best solutions [42]. GAs start with an array of feasible solutions; each expressed as a series of decision parameters. These candidate solutions are then subjected to selection, crossover, and mutation processes to generate new offspring solutions. Each resultant solution is then assessed by an objective function to determine its fitness level. Those with higher fitness are more likely to persist into subsequent generations, while those with lower fitness are phased out over time. The cycle repeats until a termination criterion is satisfied, such as reaching a predetermined number of cycles or finding an acceptable solution.

GAs achieve a delicate balance between the exploratory and exploitative aspects of optimisation [43]. Exploration involves surveying the search space to find new areas that might house superior solutions. Exploitation, on the other hand, is about improving the solutions located in promising regions. This equilibrium is realised through selection, crossover, and mutation. Selection favours the survival of fitter solutions. Crossover merges the genetic information of chosen solutions to create new offspring exhibiting a blend of characteristics. Mutation triggers random alterations in the offspring, fostering exploration by bringing unique genetic variations.

*Crossover Operator:.* The geometric crossover technique [44] has been strategically chosen for its remarkable ability to identify and uncover potential solutions that lie precisely on the edge of what can be considered a feasible solution space, as referenced in the source [17]. Moreover, this operation enables smooth transition in the search space, enhancing exploitation while preserving diversity. It is particularly beneficial for realvalued and continuous optimisation problems since it guarantees feasibility and enables convergence towards optimal regions with higher precision. Envision two parent chromosomes, represented mathematically as $A = \{a_1, a_2, ..., a_n\}$ and $B = \{b_1, b_2, ..., b_n\}$, from which the offspring are derived through a specific calculation method outlined below.

$$C = \{\sqrt{a_1 \cdot b_1}, \sqrt{a_2 \cdot b_2}, ..., \sqrt{a_n \cdot b_n}\} \tag{7}$$

$$C_i = (A_i)^\alpha \cdot (B_i)^{1-\alpha}. \tag{8}$$

In this context, the variable *i* denotes the number of individual indexes associated with each chromosome, while $\alpha$ is confined to the interval [0,1], indicating the proportion that influences the merging of the parent chromosomes. Specifically, when the value of $\alpha$ is set to $\frac{1}{2}$, thereby illustrating a balanced combination of both parent genes. Two offspring are created by swapping parent positions during the second calculation, adding variety to genetic mixing. This method also supports multiple parents, increasing genetic diversity and innovation as follows.

$$C_i = (A_i^1)^{\alpha_1}(A_i^2)^{\alpha_2}(A_i^3)^{\alpha_3}\ldots(A_i^n)^{\alpha_n}, \text{ where, } \sum_{i=1}^{n}\alpha_i = 1 \tag{9}$$

*Mutation Operator:.* A crucial mechanism in the realm of genetic algorithms is a mutation, which plays a significant role in altering one or more genes within a given population, thereby enhancing the overall variability and diversity of that population in an effort to explore the vast landscape of potential solutions more thoroughly.

To illustrate this concept, let us consider an individual represented as $A_1 = (a_1, a_2, \cdots, a_n)$, where each variable in a solution $a_i$ is confined within a specific range, defined by the lower bound $Low_b(i)$ and the upper bound $Up_b(i)$, which respectively set the limits for that variable's potential values.

A non-uniform mutation operator was used, which is designed to alter the selected variables in a manner that is not uniform across the population but rather varies depending on certain criteria. Equation 10 shows the formulation of this mutation where *iter* and *iter$_{max}$* are the current and maximum generation number, $\vartheta$ is a random number between 0 and 1, and $\beta$ is a system parameter determining the degree of non-uniformity equal to 6 in this research.

$$a_i' = \begin{cases} a_i + (Up_b(i) - a_i)\left(\vartheta \cdot \left(1 - \frac{iter}{iter_{max}}\right)\right)^\beta & if\ rand \leq \alpha \\ a_i - (a_i - Low_b(i))\left(\vartheta \cdot \left(1 - \frac{iter}{iter_{max}}\right)\right)^\beta & if\ rand > \alpha \end{cases} \tag{10}$$

*Population Size Importance:.* Population size that determines the number of solutions in it is a critical factor in determining the effectiveness of GAs. A large population promotes greater diversity and exploration but results in higher computational expense. Small populations, conversely, might converge quicker but have the potential to get stuck in suboptimal solutions. Problem complexity, search space, and computing resources can determine the selection of an ideal population size. It should measure the objectives, boundaries, and requirements specific to the problem and establish how close a solution is to global or local optimal. These fitness values are utilised by the GA to guide the search process, favouring solutions with greater fitness values. GAs can also be hybridised in a hybridisation with other optimisation methods in an attempt to enhance performance. Hybridisation strategies take advantage of the strengths of various algorithms without their weaknesses. For instance, genetic algorithms can be blended with local search techniques to enhance the performance of the genetic algorithm and convergence to improved solutions.

### 3.2.3. Single-Parent Evolutionary Algorithm

The Single-Parent evolutionary algorithm known as 1+1EA is an optimisation method [45] that begins with a starting solution, *X* and generates a new solution, *Y*, in each iteration by randomly altering one or more selected variables in *X* ($X_{iter} \in \{LB,UB\}^N$), where *UB* and *LB* represent the upper and lower bounds of the variable, respectively, and *N* denotes the number of variables. Unlike the standard 1+1EA, which employs a uniform distribution for mutation, resulting in a local search that is both non-curved and noisy, we prefer to utilise a normally distributed transformation [46]. Next, the new solution generated is evaluated and compared with its parent. If the fitness of the new solution dominates the previous one, it will be replaced. Otherwise, the new solution will be removed, and another solution generates from the parent candidate.

*Mutation Operator:.* Contrary to the default 1+1EA with a uniform random mutation, leading to non-curved and noisy search behaviour, our implementation employs a Gaussian (normally distributed) mutation scheme to enable better local search in the vicinity of the parent solution. Specifically, the mutation for each decision variable *i* is defined as:

$$Y_i = \mathcal{N}\left(\mu = X_i, \sigma^2 = 0.2 \times (UB - LB)\right) \tag{11}$$

The parameter $\mu = X_i$ defines the mean of the distribution, ensuring that mutations occur locally around the current solution. The standard deviation $\sigma$ is derived from the problem's variable bounds and is computed as $\sqrt{0.2} \times (UB - LB)$. This adaptive normal distribution ensures a more refined search around the parent solution and allows for better exploitation of promising regions in the search space. The use of Gaussian noise is particularly effective in real-valued continuous optimization problems where smooth convergence is desirable [46].

*Selection Strategy:.* Once the new solution $Y$ is generated, it undergoes fitness evaluation. The selection mechanism in 1+1EA follows a greedy strategy:

$$X^{(\text{iter}+1)} = \begin{cases} Y, & \text{if } f(Y) \leq f(X^{(\text{iter})}) \\ X^{(\text{iter})}, & \text{otherwise} \end{cases} \tag{12}$$

1+1EA offers the advantage of changing only a small number of variables in each iteration. This characteristic allows for a gradual approach towards a nearly optimal solution. However, for large-scale optimisation problems, this can incur significant costs. Empirical evidence suggests that simpler EAs can occasionally outperform more complex methods. Additionally, 1+1EA proves to be a suitable choice when the fitness function involves a combinatorial optimisation problem [47].

### 3.3. Ensemble learning models

In machine learning, ensemble models combine several different models, resulting in much better overall predictions [48]. The procedure compensates for the weaknesses that may result from overfitting or bias. Now, this can be looked at broadly under three sections: first, bagging; second, boosting; third, stacking. As the model trains on varied subsets to reduce variance, an exemplary model created with the procedure of '*bagging*' is a Random Forest [49]. Boosting, in a manner similar to that of Adaptive Boosting (AdaBoost) [50], XGBoost [51] and Gradient Boosting [52], tries to decrease the bias by iteratively correcting the model's past mistakes. Stacking takes an approach to meta-learning by making use of a high-order model to combine the predictions of lower-level base learners. Through their ability to aggregate diverse models, these ensemble methods have been successful at generalising and providing performances when single-model techniques failed.

Among others, three significant advantages, which can be identified as more important than the benefits created by more traditional ML methods in this work, are enhanced accuracy, robustness, and adaptability. This will generally lead to better overall performances since the strengths of ensemble methods aggregate several models together to minimise both bias and variance toward errors [53]. It is much more robust towards noise and outliers among the data points. Above all, most of the ensemble methods can be seamlessly integrated with almost all data and problem types, either classification or regression, natural selections for complex practical applications—advantages that fully implement their valuable contribution to achieving state-of-the-art machine learning tasks.

### 3.3.1. Stacking ensemble models

Stacking ensemble models typically combines a set of base models—usually referred to as the level-0 learners—predictions via a higher-level meta-model, commonly referred to as the level-1 learner, for improving results [54]. Each of the base models uses different algorithms in their training with the same dataset with the aim of ensuring diversity that would utilise each of their unique strengths. It is crucial that the meta-model learn how to effectively combine output from these base learners in a refined and generally more fitting final prediction [55]. The key insight to stacking is when different models specialise in combining strengths and other aspects of the problem. The advantages include increased predictive accuracy arising due to the combination of various models and immense versatility regarding the handling of complex challenges.

The framework of stacking can be described in the following steps [48]: The base models are trained using certain algorithms, the choice of which depends on the problem domain and requirements of the user. This step involves preparing the base learners using the provided training data. These, in turn, are used to develop a new dataset. This new dataset will contain the predicted outputs of the base models as new features and the actual target labels as corresponding target values. For example, any instance in the original dataset *R* if of the form $a_i, f(a)_i$, then the same instance in the new dataset created will be in the form $\hat{a}_i, f(a)_i$ where $\hat{a}_i$ is composed of the various outputs $h_1(a_i), h_2(a_2),...,h_T(a_i)$ from the different base models. The meta-learner is then trained using this new dataset, hence learning how to integrate the predictions of the base models [48]. The meta-model is then deployed to combine the outputs from the base models for new, unseen data. In stacking, for an out-of-sample instance *a*, the ultimate prediction is a function from the meta-learner: $\hat{h}(h_1(a), h_2(a),...,h_T(a))$, with respect to outputs from the base models—the level-0 models. However, despite its potential for high accuracy, stacking is not as widely adopted as either bagging or boosting due to the complexity of implementation and possible data leakage if not treated appropriately.

*3.3.2. Bagging ensemble models*

Bagging, short for bootstrap aggregating, is an ensemble technique aimed at reducing the variance of model predictions and improving generalisation by combining multiple models [56]. These models are trained independently on diverse, randomly generated subsets of either the training data or input features. Each is trained separately on a different, random subset of the training data or input features. Bootstrapping refers to creating *M* sets of data $\{D_1, D_2,...,D_M\}$ with size *n*, each drawn with replacement from the original training set *D*. Mathematically, for each dataset $D_m$, with *m* = 1,2,...,*M*, we have:

$$D_m = \{(x_i, y_i)\}_{i=1}^n, \quad D_m \sim D \tag{13}$$

Each subset $D_m$ is used to train a base model $h_m(x)$. The final prediction is made by aggregating the outputs of these base models: For regression tasks, the prediction is given by the average:

$$\hat{y} = \frac{1}{M} \sum_{m=1}^{M} h_m(x) \tag{14}$$

The final prediction is made by aggregating the outputs of these models, using majority voting for classification tasks or averaging for regression tasks [57]. A prominent application of Bagging is the Random Forest algorithm, which builds numerous decision trees and combines their results to produce stable and accurate predictions.

Relative to stacking and boosting, Bagging possesses distinct advantages. Unlike boosting, which sequentially trains models with the emphasis being placed on rectifying errors from the previous iterations, Bagging trains its base models in parallel and independently from one another [58]. The parallel approach reduces the risk of overfitting and enhances computational efficiency. In addition, while stacking combines the heterogeneous algorithm predictions using a meta-learner, Bagging tends to employ a single algorithm type to create homogeneous models, which are simpler to implement. Another significant benefit of Bagging is that it is robust to noisy data and outliers because boosting does not assign extra weight to difficult instances. Bagging is particularly valuable in applications where variance reduction and generating consistent, generalised predictions are key goals.

*3.3.3. Voting ensemble models*

Voting is one of the most straightforward ensemble learning techniques, and the whole perspective is that combining predictions from many models results in overall improvements in performance. This approach works by aggregating base model outputs by majority vote or averaging [59].

Voting ensembles can be composed of homogeneous models (i.e., models of the same type trained on different data subsets) or heterogeneous models (i.e., models based on different algorithms). There are two main types of voting: majority voting for classification tasks and averaging for regression tasks. In an $N_C$ class in a classification problem with $N_e$ base classifiers, the output of the $i^{th}$ classifier for class *c* is denoted as $O_{i,c} \in$

{0,1}, where $O_{i,c}$ = 1 if the classifier $h_i$ predicts class $c_r$ and $O_{i,c}$ = 0 otherwise. With majority voting, the ensemble prediction $\omega_{c^*}$ is the class label that receives the most votes:

$$c^* = \arg \max_{c \in \{1,\ldots,N_C\}} \sum_{i=1}^{N_e} O_{i,c} \tag{15}$$

In weighted majority voting, every classifier $h_i$ is assigned a weight $w_i$, which is its estimated reliability or generalisation ability. The class $c^*$ is predicted by computing the weighted sum of votes across all classifiers:

$$c^* = \arg \max_{c \in \{1,\ldots,N_C\}} \sum_{i=1}^{N_e} w_i \cdot O_{i,c} \tag{16}$$

For regression, voting is replaced by averaging. Each base model produces a real-valued output $h_i(x)$, and the final prediction $\hat{y}$ is taken to be the average (or weighted average) of all base outputs:

$$\hat{y} = \frac{1}{N_e} \sum_{i=1}^{N_e} h_i(x) \text{ (unweighted)} \quad \text{or} \quad \hat{y} = \sum_{i=1}^{N_e} w_i \cdot h_i(x) \text{ (weighted)} \tag{17}$$

This ensemble process is simple yet effective, particularly when the base learners are heterogenous because it tends to reduce variance while enhancing robustness.

*3.3.4. Boosting Ensemble models*

Extreme gradient boosting XGBoost is an innovative machine-learning methodology that enhances treebased models through an assembly of classification and regression trees (CART) [51]. This methodology is structured on a gradient-boosting framework, which enables simultaneous tree boosting. The tree assembly model merges numerous weak learners to forecast the output by applying an incremental training approach. The steps of this incremental training are as follows: initially, the full scope of input data is adjusted by the first learner, after which the residuals, which are used to rectify the deficiencies of a weak learner, are modified by a subsequent learner. This adjustment procedure is repeated multiple times until the termination condition is met. The final prediction of the model is then derived as the cumulative prediction of all learners. The parallel procedures are autonomously executed during the training phase, thereby facilitating the efficient use of computational resources [60]. Moreover, in order to deal with over-fitting issues, an advanced regularised formulation is applied as follows:

$$L(\omega) = \sum_i^N d(y_i', y_i) + \sum_k \lambda(f_k) \tag{18}$$

where $d$ plays the role of the loss function to calculate the difference between the predicted value and true value. $\lambda$ is the regularisation function to penalise the ld complexity of the model. $\alpha$ is a threshold to extend the leaf node. The weight of the leaf and regularisation parameter are shown by $s$ and $\beta$, and $T$ is the number of tree leaves.

XGBoost offers several advantages contributing to its widespread adoption and success in various domains. It can be used in a wide range of data types, including numerical, categorical, and text data. Additionally, XGBoost allows customising loss functions, enabling users to specify their objective functions and tailor the model to distinct conditions. Another benefit of XGBoost is offering valuable information about the significance of features, qualifying the users to comprehend how different predictors contribute to the model's overall performance [61]. Assessing feature importance simplifies the identification of influential variables, facilitates feature selection, and enhances the understanding of the underlying data.

*3.4. Proposed Adaptive Evolutionary Ensemble learning model*

This section outlines the technical aspects of the proposed neuro-evolutionary model for forecasting energy consumption in smart buildings. The methodology comprises six main steps: baseline model comparison, hyper-parameter selection, and optimisation of the chosen model.

- Initially, we selected 15 diverse ML models for evaluation, including four traditional algorithms: Support Vector Machine (SVM), Logistic Regression (LR), Bayesian Linear Regression (BR), and k-nearest Neighbors (KNN). Additionally, we incorporated three neural network architectures: Multi-Layer Perceptron (MLP), Dense Neural Network (DNN), and Convolutional Deep Neural Network (CDNN).
  To further enhance diversity, three tree-based models, RF [62], DT, and Extra Tree (ET) were included. Lastly, seven ensemble models were trained and assessed: XGBoost [51], AdaBoost [63], Gradient Boosting Regressor (GBR) [64], Histogram-Based Gradient Boosting Regressor (HGBR) [65], Categorical Boosting (CatBoost) [66], and Light Gradient Boosting Machine (LGBM). The specific configurations [52] used for training these models are detailed in Table S3.

- We developed a robust hybrid ensemble framework that incorporates three strategies: stacking, bagging, and voting, to enhance the learning capability of an individual model by improving its predictive accuracy. This effectively fuses the strengths of each method in leveraging their complementary mechanisms toward a more accurate and reliable predictive model.

- In the stacking ensemble model, the best-performing model among 15 candidates was selected as the initial base learner, with linear regression as the meta-learner. Additional base learners were identified using a greedy search approach, incrementally adding models that improved performance metrics such as accuracy or error reduction. At each step, the combination of base learners yielding the highest performance was retained, ensuring the inclusion of only the most effective models while avoiding redundancy. The same technique was applied to optimise the meta-learner, further enhancing the ensemble's predictive capability (See Figure 8). The details of the stacking ensemble model procedure can be seen in Algorithm 1 (In Appendix).

- In the second proposed ensemble model, we began with six superior-performing ML methods embedded in a weighted majority vote framework. Then, the models went one by one into the removing process, and the performance of the resultant ensemble was re-evaluated in the absence of the model that was being removed. The process was reiterated to see if this improved the accuracy of the prediction result. Then, weights within the final resultant ensemble were also optimised using the Nelder-Mead local search. The result was an optimal voting model, which only contained two methods, XGB and LGBM, having equal weights (See Table 7 and Figure 12).

- Leveraging the unique advantages of bagging ensemble models—such as reducing variance, preventing overfitting, and improving stability—we developed an adaptive bagging framework. This approach involved evaluating nine models trained and tested within the bagging framework. The best-performing model, Extra Trees, was then selected for further optimisation. To enhance its performance, we applied a fast and robust optimisation algorithm, 1+1 Evolutionary Algorithm (1+1EA), to fine-tune its hyperparameters, ensuring optimal predictive accuracy and efficiency (See Table6 and Figure 11). • Finally, we implemented and compared four widely used meta-heuristic algorithms— GA, DE, Particle Swarm Optimisation (PSO), and 1+1EA—to optimise the hyper-parameters of the proposed ensemble models, assessing their effectiveness and performance. Meta-heuristic algorithms explore and find the optimal and feasible combination of parameters for maximising the prediction accuracy of total power consumption using IoT-based collected information. The formulation is represented as follows.

$$\lambda(f) = \alpha T + 1/2\beta \|s\|^2 \qquad (19)$$

where $\Psi$ and $N_h$ are the search space and number of hyper-parameters listed in Table S2. $f(h)$ evaluates the machine learning effectiveness with the set of hyper-parameters $h$ that should be maximised. The fitness function ($f(h)$) is subjected to the boundary constraints ($\Lambda$) listed in Table S3.

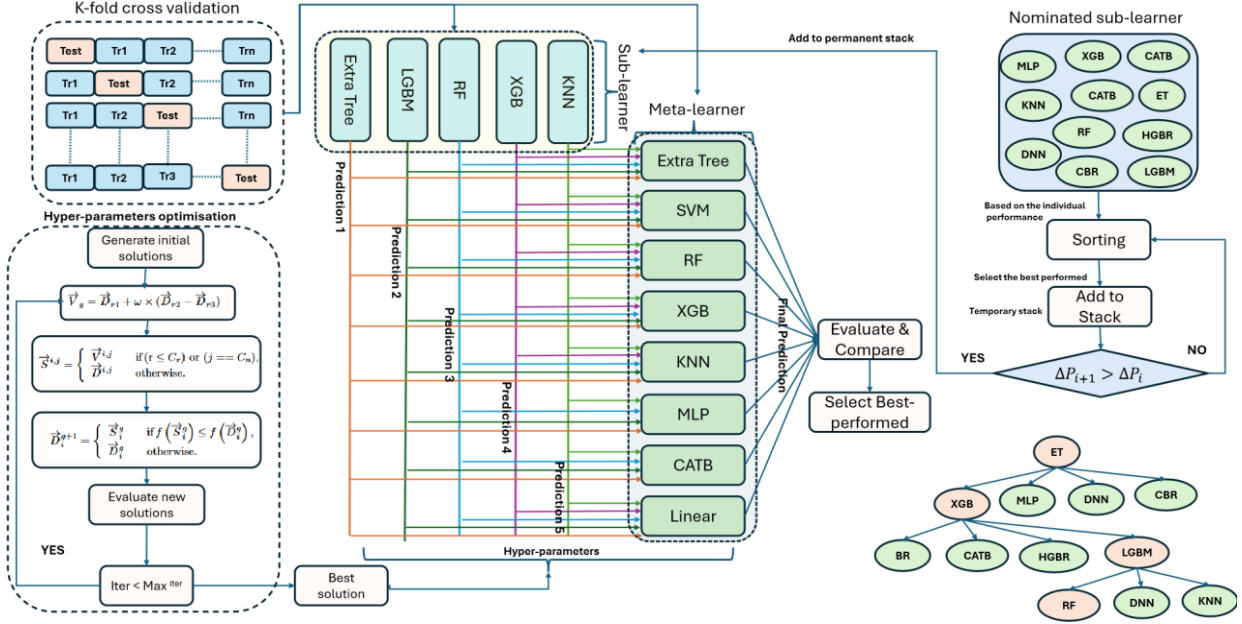

Figure 8: Schematic flowchart illustrating the workflow of the proposed adaptive evolutionary stacking ensemble model, highlighting the ensemble tree model's performance as determined by the greedy search method.

For the stacking ensembles' meta-learner, we selected the top ten models performing on cross-validation metrics (R-value, MAE, RMSE). This ensured that only those models with very high individual predictive ability were chosen for the second-level learning process. To construct the sub-learner block in stacking and voting ensembles, we employed a greedy forward selection strategy. This strategy begins with the top-performing model and gradually includes the subsequent candidates one by one, only retaining a model if its addition leads to a gain in average performance for all measures of evaluation. The procedure is iterated until no more models can further enhance the predictive performance of the ensemble. Using this method, we prepared and tested ten stacked scenarios, each being compared in terms of performance gains. Similarly, in the case of bagging ensembles, we created eight models with the ensemble of the top-performing individual learners under a single feature space. In the case of voting ensembles, a greedy selection strategy demonstrated that gains in performance plateaued after two base models at maximum, so we had six finetuned voting models. Such systematic selection also ensures that resultant ensemble structures are not only high-performing but also efficient in computation and non-redundant.

In order to ensure guarantees of convergence and stability of individual learners in the ensemble, our proposed framework (Figure 8) contains several precautions designed to mitigate the impact of non-converging models on the overall process of training. Each candidate learner is first independently tested with Kfold cross-validation, separating any instability or non-convergence to that specific model so that it cannot contaminate the integrity of the ensemble. Suppose a learner fails to converge or has a score below some threshold. In that case, the greedy selection strategy, illustrated on the right of the schematic, removes it systematically from the stacking structure. This is based on the difference in performance (ΔP), and only those sub-learners that enhance the ensemble's overall predictive accuracy are retained in the transient and, subsequently, the permanent stack. Moreover, the hyperparameter optimisation module (in the top centre of the figure) enhances convergence likelihood through the application of a metaheuristic search strategy to incrementally tune each learner's parameters adaptively. This serves the purpose of bypassing local optimum areas of parameter space, which else could induce training instability or divergence. Finally, the meta-learner is trained only after the sub-learner block has been completed from converged and validated models. Therefore, any non-converging learner is naturally excluded from the final ensemble, and the pipeline for training is stable, robust, and driven by validated performance improvement.

---

**Algorithm 1** Adaptive Evolutionary Stacking learning model

---
1: **procedure** EA STACKING
2:   **Initialise** Define $P_{sub} = \{S_1, S_2, ..., S_{N_e}\}$, $N_e$, $Min_{N_e}$    ▷ Base ML algorithm and number of base learners
3:   Define $N_{pop}$, $CR$, $F$    ▷ DE settings, population size, crossover and mutation rate
4:   Training data $P = (x_{s1}, y_{s1}), (x_{s2}, y_{s2}), \ldots, (x_{sn}, y_{sn})$
5:   **while** $i_{sub} \leq N_e$ **do**
6:     Train $i_{sub}$ base learning model    ▷ Step 1
7:     Fit a base learner $S_i$ by P
8:     **if** $\Delta ACC_{i_{sub}} > 0$ **then**    ▷ If the Stack model performance improved by $S_i$
9:       Add base learner $S_i$ to Stack
10:       Update $\Delta ACC$
11:     **end if**
12:   **end while**
13:   Generate a new set from P    ▷ Step 2
14:   **while** $i_{sub} \leq N_e$ **do**
15:     $\hat{P}\{\hat{x}_i, y_i\}$, where $\hat{x}_i = \{S_1(x_1), S_2(x_2), \ldots, S_{N_e}(x_i)\}$
16:   **end while**
17:   Train and test the meta-learners by $\hat{P}$    ▷ Step 3
18:   **while** $j \leq Max_{iter}$ **do**    ▷ Run hyper-parameters optimisation
19:     Generate initial population of random solution    ▷ Step 4
20:     Select 3 random solutions, $r_1, r_2, r_3 \in (1, N_{pop})$, with $r_1 \neq r_2 \neq r_3 \neq j$
21:     **while** $k \leq N_{pop}$ **do**
22:       $\vec{V}_g = \vec{D}_{r1} + \omega \times (\vec{D}_{r2} - \vec{D}_{r3})$    ▷ Mutation operator
23:       $\vec{S}^{i,j} = \begin{cases} \vec{V}^{i,j} & \text{if}(r \leq C_r) \text{ or } (j == C_n), \\ \vec{D}^{i,j} & \text{otherwise.} \end{cases}$    $j = 1, 2, \ldots N_D$    ▷ Crossover operator
24:       $\vec{D}_i^{g+1} = \begin{cases} \vec{S}_i^g & \text{if } f(\vec{S}_i^g) \leq f(\vec{D}_i^g), \\ \vec{D}_i^g & \text{otherwise.} \end{cases}$    ▷ Selection better solution
25:     **end while**
26:     $Best_s = Max(f(D^g))$    ▷ Select the best solution
27:   **end while**
28:   Train and test the Stack model using $Best_s$
29: **end procedure**

---

## 4. Experimental results

This study presents the outcomes achieved through the utilisation of the proposed three hybrid evolutionary ensemble strategies and 15 popular ML models in predicting the total power consumption of appliances based on a hybrid dataset of meteorological parameters, energy use of appliances, temperature, humidity, and lighting energy consumption of different sections collected by 18 sensors in a building which is located in Stambruges, Mons in Belgium. In addition, a concise analysis of the key discoveries from this research is provided. With regard to developing a comprehensive and robust comparative prediction framework, 14 effective ML models were selected. Each model was independently trained ten times based on 10-fold crossvalidation, and the percentage of training, validating, and testing were 80%, 10%, and 10%, respectively. We employed a parallelised K-fold cross-validation strategy to address computational demands associated with training advanced ensemble models through K-fold cross-validation. Because every fold in cross-validation is independent, model training and validation for every fold were executed in parallel on multiple CPU cores. This significantly reduced overall runtime without sacrificing cross-validation's strengths in robustness and generalisability. Specifically, we utilised parallel computing abilities in Python's scikit-learn package (via n jobs=-1) and tuned our model pipelines to allow parallel processing without compromising reproducibility.

*4.1. Evaluation metrics*

To assess the performance of the proposed hybrid models alongside the other 15 ML models, we utilised seven widely recognized evaluation metrics [67], as outlined in Table 3. Among these, MSE, RMSE, MAE, MSLE, and SMAPE are metrics where lower values indicate better performance. Conversely, higher values are more desirable for EVS and R-value, as they reflect greater predictive accuracy and a stronger linear relationship between predictions and true values. Where $N_s$ represents the total number of samples, $f_e(k)$ denotes the estimated (predicted) output of the model for the $k^{th}$ sample, and $f_t(k)$ is the corresponding true (target) value.

Table 3: The performance evaluation metrics for the prediction models

| Metrics | Definition | Equation |
|---|---|---|
| MSE | Mean square error | $MSE = \frac{1}{N_s}\sum_{k=1}^{N_s}(f_e(k)-f_t(k))^2$ |
| RMSE | Root mean square error | $RMSE = \sqrt{\frac{1}{N_s}\sum_{k=1}^{N_s}(f_e(k)-f_t(k))^2}$ |
| MAE | Mean absolute error | $MAE = \frac{1}{N_s}\sum_{k=1}^{N_s}|f_e(k)-f_t(k)|$ |
| MSLE | Mean squared log error | $MSLE = \frac{1}{N_s}\sum_{k=1}^{N_s}(log_e(1+f_t(k)) - log_e(1+f_e(k)))^2$ |
| SMAPE | Symmetric mean absolute percentage error | $SMAPE = \frac{1}{N_s}\sum_{k=1}^{N_s}\frac{|f_t(k)-f_e(k)|}{(\frac{1}{2}(f_t(k)+f_e(k)))} \times 100$ |
| EVS | Explained variance score | $EVS = 1 - \frac{\sum_{k=1}^{s} Variance(f_t(k)-f_e(k))}{Variance(f_t(k))}$ |
| R-value | Pearson correlation coefficient | $R = \frac{\frac{1}{N_s}\sum_{k=1}^{s}(f_e(k)-\overline{f_e})(f_t(k)-\overline{f_t})}{\sqrt{\frac{1}{N_s}\sum_{k=1}^{N_s}(f_e(k)-\overline{f_e})^2} \times \sqrt{\frac{1}{N_s}\sum_{k=1}^{N_s}(f_e(k)-\overline{f_t})^2}}$ |

*4.2. Quantitative Evaluation and Statistical Analysis*

This section provides a detailed quantitative comparison of the proposed models on the basis of statistical performance metrics. Certain error measures and R-values are used to compare the accuracy, robustness, and generalisation capacity of the models with various experimental configurations. Comparative statistical analysis with conventional methods is also included to reasonably validate the excellence of the proposed framework in predicting energy consumption in smart buildings.

*4.2.1. Baseline models experimental results*

Table 4 shows the statistical results corresponding to 14 ML models' performance to predict the power appliances' consumption using six evaluation metrics. The analysis of the provided Table 4 reveals intriguing findings regarding various models' prediction accuracy (R-value). Notably, the XGBoost model emerged as the top performer, exhibiting an impressive average accuracy of 73% across ten runs. We can see that in the best-case scenario, this model achieved a remarkable accuracy of 75%. Furthermore, the GBM, CatBoost, and EBM models also demonstrated considerable levels of accuracy, with respective values of 72.8%, 72.3%, and 71.4%. It is worth mentioning that, in general, the performance of neural networks and deep learning models, such as Dense (DNN) and convolutional (CDNN) deep models, fell slightly behind ensemble models in terms of average accuracy. However, it is noteworthy to investigate that the AdaBoost model proved to be an oddity to this trend. These findings shed light on the comparative performance of different models, providing valuable insights for future analysis and decision-making processes.

Table 4: Statistical analysis results of the appliances power consumption prediction using 14 well-known machine learning methods, neural networks, deep learning, ensemble, tree-based and hybrid methods.

|  | SVM | | | | | | | MLP | | | | | |
|---|---|---|---|---|---|---|---|---|---|---|---|---|---|
|  | RMSE | MAE | MSLE | SMAPE | EVS | R-value |  | RMSE | MAE | MSLE | SMAPE | EVS | R-value |
| Min | 9.45E+01 | 4.11E+01 | 2.99E-01 | 3.28E+01 | 9.46E-02 | 3.72E-01 | Min | 8.19E+01 | 4.24E+01 | 2.43E-01 | 3.46E+01 | 2.79E-01 | 5.29E-01 |
| Max | 1.05E+02 | 4.47E+01 | 3.30E-01 | 3.49E+01 | 1.10E-01 | 4.15E-01 | Max | 8.83E+01 | 5.46E+01 | 4.60E-01 | 4.71E+01 | 3.59E-01 | 6.00E-01 |
| Mean | 1.01E+02 | 4.30E+01 | 3.14E-01 | 3.37E+01 | 1.01E-01 | 3.85E-01 | Mean | 8.55E+01 | 4.69E+01 | 3.26E-01 | 4.02E+01 | 3.20E-01 | 5.67E-01 |
| Median | 1.02E+02 | 4.32E+01 | 3.15E-01 | 3.36E+01 | 1.00E-01 | 3.81E-01 | Median | 8.56E+01 | 4.67E+01 | 3.05E-01 | 4.06E+01 | 3.19E-01 | 5.66E-01 |
| STD | 2.61E+00 | 1.09E+00 | 9.25E-03 | 5.24E-01 | 4.36E-03 | 1.16E-02 | STD | 2.16E+00 | 3.05E+00 | 6.57E-02 | 3.05E+00 | 2.15E-02 | 1.86E-02 |
|  | DNN | | | | | | | CDNN | | | | | |
|  | RMSE | MAE | MSLE | SMAPE | EVS | R-value |  | RMSE | MAE | MSLE | SMAPE | EVS | R-value |
| Min | 7.22E+01 | 3.72E+01 | 2.04E-01 | 3.10E+01 | 3.17E-01 | 5.82E-01 | Min | 7.38E+01 | 3.33E+01 | 1.77E-01 | 2.68E+01 | 3.56E-01 | 6.35E-01 |
| Max | 8.55E+01 | 4.37E+01 | 2.38E-01 | 3.47E+01 | 4.80E-01 | 7.13E-01 | Max | 8.29E+01 | 3.66E+01 | 1.97E-01 | 2.80E+01 | 4.53E-01 | 6.91E-01 |
| Mean | 8.17E+01 | 4.08E+01 | 2.23E-01 | 3.30E+01 | 3.89E-01 | 6.43E-01 | Mean | 7.80E+01 | 3.48E+01 | 1.87E-01 | 2.72E+01 | 4.01E-01 | 6.59E-01 |
| Median | 8.25E+01 | 4.15E+01 | 2.21E-01 | 3.34E+01 | 3.77E-01 | 6.38E-01 | Median | 7.77E+01 | 3.49E+01 | 1.87E-01 | 2.72E+01 | 3.96E-01 | 6.56E-01 |
| STD | 4.06E+00 | 1.99E+00 | 1.29E-02 | 1.32E+00 | 4.57E-02 | 3.58E-02 | STD | 2.71E+00 | 1.04E+00 | 7.20E-03 | 3.95E-01 | 2.95E-02 | 1.70E-02 |
|  | HGBR | | | | | | | DT | | | | | |
|  | RMSE | MAE | MSLE | SMAPE | EVS | R-value |  | RMSE | MAE | MSLE | SMAPE | EVS | R-value |
| Min | 7.11E+01 | 3.61E+01 | 1.81E-01 | 3.02E+01 | 4.26E-01 | 6.54E-01 | Min | 8.56E+01 | 3.56E+01 | 2.13E-01 | 2.60E+01 | 1.54E-01 | 5.85E-01 |
| Max | 7.93E+01 | 3.96E+01 | 2.05E-01 | 3.17E+01 | 5.12E-01 | 7.23E-01 | Max | 9.50E+01 | 3.94E+01 | 2.43E-01 | 2.81E+01 | 3.05E-01 | 6.46E-01 |
| Mean | 7.53E+01 | 3.76E+01 | 1.94E-01 | 3.10E+01 | 4.64E-01 | 6.84E-01 | Mean | 9.01E+01 | 3.77E+01 | 2.23E-01 | 2.70E+01 | 2.27E-01 | 6.17E-01 |
| Median | 7.57E+01 | 3.78E+01 | 1.95E-01 | 3.09E+01 | 4.61E-01 | 6.82E-01 | Median | 9.03E+01 | 3.77E+01 | 2.22E-01 | 2.70E+01 | 2.37E-01 | 6.17E-01 |
| STD | 2.31E+00 | 8.54E-01 | 6.40E-03 | 3.88E-01 | 2.18E-02 | 1.73E-02 | STD | 2.38E+00 | 9.20E-01 | 8.54E-03 | 4.24E-01 | 4.64E-02 | 2.01E-02 |
|  | EBM | | | | | | | XGB | | | | | |
|  | RMSE | MAE | MSLE | SMAPE | EVS | R-value |  | RMSE | MAE | MSLE | SMAPE | EVS | R-value |
| Min | 6.61E+01 | 3.16E+01 | 1.57E-01 | 2.55E+01 | 4.58E-01 | 6.87E-01 | Min | 6.76E+01 | 3.15E+01 | 1.48E-01 | 2.46E+01 | 4.84E-01 | 7.03E-01 |
| Max | 7.74E+01 | 3.54E+01 | 1.75E-01 | 2.70E+01 | 5.47E-01 | 7.41E-01 | Max | 7.52E+01 | 3.48E+01 | 1.67E-01 | 2.59E+01 | 5.58E-01 | 7.49E-01 |
| Mean | 7.15E+01 | 3.37E+01 | 1.64E-01 | 2.62E+01 | 5.06E-01 | 7.14E-01 | Mean | 7.09E+01 | 3.27E+01 | 1.55E-01 | 2.52E+01 | 5.28E-01 | 7.30E-01 |
| Median | 7.11E+01 | 3.36E+01 | 1.64E-01 | 2.62E+01 | 5.06E-01 | 7.13E-01 | Median | 7.09E+01 | 3.25E+01 | 1.54E-01 | 2.52E+01 | 5.27E-01 | 7.30E-01 |
| STD | 3.02E+00 | 9.74E-01 | 5.03E-03 | 4.00E-01 | 2.35E-02 | 1.53E-02 | STD | 1.93E+00 | 7.34E-01 | 5.23E-03 | 3.80E-01 | 1.97E-02 | 1.24E-02 |
|  | AdaB | | | | | | | CatB | | | | | |
|  | RMSE | MAE | MSLE | SMAPE | EVS | R-value |  | RMSE | MAE | MSLE | SMAPE | EVS | R-value |
| Min | 1.00E+02 | 6.62E+01 | 5.12E-01 | 5.16E+01 | -2.83E-01 | 3.46E-01 | Min | 6.85E+01 | 3.43E+01 | 1.71E-01 | 2.92E+01 | 4.79E-01 | 6.94E-01 |
| Max | 1.82E+02 | 1.67E+02 | 1.68E+00 | 1.03E+02 | 1.02E-01 | 4.01E-01 | Max | 7.48E+01 | 3.64E+01 | 1.87E-01 | 3.06E+01 | 5.52E-01 | 7.45E-01 |
| Mean | 1.31E+02 | 1.02E+02 | 8.96E-01 | 6.88E+01 | -7.98E-02 | 3.69E-01 | Mean | 7.10E+01 | 3.55E+01 | 1.75E-01 | 2.98E+01 | 5.21E-01 | 7.23E-01 |
| Median | 1.26E+02 | 9.56E+01 | 8.26E-01 | 6.62E+01 | -7.00E-02 | 3.66E-01 | Median | 7.07E+01 | 3.56E+01 | 1.75E-01 | 2.98E+01 | 5.20E-01 | 7.24E-01 |
| STD | 2.04E+01 | 2.40E+01 | 2.76E-01 | 1.21E+01 | 9.37E-02 | 1.40E-02 | STD | 1.72E+00 | 6.65E-01 | 3.98E-03 | 3.08E-01 | 1.61E-02 | 1.17E-02 |
|  | BR | | | | | | | RF | | | | | |
|  | RMSE | MAE | MSLE | SMAPE | EVS | R-value |  | RMSE | MAE | MSLE | SMAPE | EVS | R-value |
| Min | 9.00E+01 | 5.20E+01 | 3.73E-01 | 4.57E+01 | 1.52E-01 | 3.90E-01 | Min | 6.83E+01 | 3.36E+01 | 1.64E-01 | 2.78E+01 | 4.67E-01 | 6.84E-01 |
| Max | 1.00E+02 | 5.48E+01 | 4.34E-01 | 4.80E+01 | 1.88E-01 | 4.35E-01 | Max | 7.61E+01 | 3.69E+01 | 1.82E-01 | 2.94E+01 | 5.26E-01 | 7.34E-01 |
| Mean | 9.46E+01 | 5.32E+01 | 3.90E-01 | 4.67E+01 | 1.66E-01 | 4.07E-01 | Mean | 7.20E+01 | 3.49E+01 | 1.74E-01 | 2.85E+01 | 4.96E-01 | 7.07E-01 |
| Median | 9.46E+01 | 5.30E+01 | 3.86E-01 | 4.66E+01 | 1.64E-01 | 4.04E-01 | Median | 7.18E+01 | 3.47E+01 | 1.75E-01 | 2.86E+01 | 4.97E-01 | 7.08E-01 |
| STD | 2.52E+00 | 7.13E-01 | 1.54E-02 | 5.37E-01 | 1.23E-02 | 1.54E-02 | STD | 2.42E+00 | 8.78E-01 | 5.55E-03 | 4.10E-01 | 1.99E-02 | 1.55E-02 |
|  | GBM | | | | | | | LGBM | | | | | |
|  | RMSE | MAE | MSLE | SMAPE | EVS | R-value |  | RMSE | MAE | MSLE | SMAPE | EVS | R-value |

| | | | | | | | | | | | | |
|---|---|---|---|---|---|---|---|---|---|---|---|---|
| Min | 6.80E+01 | 3.28E+01 | 1.57E-01 | 2.72E+01 | 4.98E-01 | 7.06E-01 | Min | 7.71E+01 | 4.25E+01 | 2.60E-01 | 3.89E+01 | 3.13E-01 | 6.16E-01 |
| Max | 7.40E+01 | 3.55E+01 | 1.73E-01 | 2.85E+01 | 5.53E-01 | 7.44E-01 | Max | 8.68E+01 | 4.55E+01 | 2.81E-01 | 4.06E+01 | 3.59E-01 | 6.78E-01 |
| Mean | 7.08E+01 | 3.42E+01 | 1.66E-01 | 2.80E+01 | 5.28E-01 | 7.28E-01 | Mean | 8.33E+01 | 4.43E+01 | 2.72E-01 | 4.00E+01 | 3.35E-01 | 6.47E-01 |
| Median | 7.02E+01 | 3.42E+01 | 1.67E-01 | 2.80E+01 | 5.31E-01 | 7.30E-01 | Median | 8.40E+01 | 4.45E+01 | 2.72E-01 | 4.01E+01 | 3.37E-01 | 6.48E-01 |
| STD | 1.93E+00 | 9.19E-01 | 4.96E-03 | 3.74E-01 | 1.45E-02 | 1.07E-02 | STD | 2.81E+00 | 8.44E-01 | 4.72E-03 | 4.12E-01 | 1.07E-02 | 1.48E-02 |

*4.2.2. Ensemble learning models result*

In this section, we present a detailed discussion, analysis, and comparison of the performance of the three proposed evolutionary ensemble models: stacking, bagging, and voting. Finally, we evaluate these strategies against each other and identify the best-performing approach, providing recommendations based on the results.

*Stacking ensemble models finding.* We evaluated the performance of various stacking models by combining multiple ML models as base learners and integrating them with meta-learners, as detailed in Table 5. The highest average accuracy, 80.3%, was achieved with a combination of ExtraTree, LGBM, RF, and KNN as base learners, paired with meta-learners such as Linear Regression or MLP, both yielding similar results. On average, stacking models demonstrated approximately a 10% improvement in prediction accuracy compared to individual ML models. Regarding MAE, the stacking model comprising ExtraTree, LGBM, RF, and KNN with Linear Regression as the meta-learner outperformed XGB, LGBM, and RF, with improvements of 91.2%, 159.0%, and 102.3%, respectively.

Table 5: Statistical analysis results of the appliances power consumption prediction using 10 Stacking ensemble methods.

| Stacking (ExtraTree+LGBM+RF+KNN/Cat) | | | | | | | Stacking(ExtraTree+LGBM+RF+KNN/linear) | | | | | | |
|---|---|---|---|---|---|---|---|---|---|---|---|---|---|
| Metric | RMSE | MAE | R value | MSLE | EVS | SMAPE | Metric | RMSE | MAE | R value | MSLE | EVS | SMAPE |
| Min | 3.04E+01 | 1.62E+01 | 7.58E-01 | 6.96E-02 | 5.56E-01 | 1.88E+01 | Min | 2.70E+01 | 1.51E+01 | 7.53E-01 | 6.63E-02 | 5.67E-01 | 1.84E+01 |
| Max | 3.74E+01 | 1.91E+01 | 8.29E-01 | 9.86E-02 | 6.86E-01 | 2.08E+01 | Max | 3.75E+01 | 1.89E+01 | 8.45E-01 | 9.20E-02 | 7.10E-01 | 2.08E+01 |
| Mean | 3.45E+01 | 1.78E+01 | 7.91E-01 | 8.46E-02 | 6.22E-01 | 1.95E+01 | Mean | 3.29E+01 | 1.71E+01 | **8.03E-01** | 8.02E-02 | 6.44E-01 | 1.93E+01 |
| Median | 3.46E+01 | 1.78E+01 | 7.85E-01 | 8.46E-02 | 6.16E-01 | 1.95E+01 | Median | 3.36E+01 | 1.73E+01 | 8.06E-01 | 8.05E-02 | 6.47E-01 | 1.94E+01 |
| STD | 2.04E+00 | 8.05E-01 | 2.03E-02 | 6.98E-03 | 3.42E-02 | 5.45E-01 | STD | 2.62E+00 | 9.82E-01 | 2.10E-02 | 6.78E-03 | 3.33E-02 | 5.94E-01 |
| Stacking(ExtraTree+LGBM+RF+KNN/MLP) | | | | | | | Stacking(ExtraTree+LGBM+RF+KNN+XGB/CBR) | | | | | | |
| Metric | RMSE | MAE | R value | MSLE | EVS | SMAPE | Metric | RMSE | MAE | R value | MSLE | EVS | SMAPE |
| Min | 3.05E+01 | 1.56E+01 | 7.35E-01 | 6.23E-02 | 5.39E-01 | 1.74E+01 | Min | 3.20E+01 | 1.69E+01 | 7.41E-01 | 7.71E-02 | 5.42E-01 | 1.85E+01 |
| Max | 3.82E+01 | 1.98E+01 | 8.40E-01 | 9.68E-02 | 6.98E-01 | 2.14E+01 | Max | 3.87E+01 | 1.99E+01 | 8.32E-01 | 1.01E-01 | 6.87E-01 | 2.11E+01 |
| Mean | 3.38E+01 | 1.75E+01 | **8.03E-01** | 8.24E-02 | 6.43E-01 | 1.95E+01 | Mean | 3.47E+01 | 1.81E+01 | 7.88E-01 | 8.62E-02 | 6.17E-01 | 1.97E+01 |
| Median | 3.33E+01 | 1.75E+01 | 8.08E-01 | 8.33E-02 | 6.49E-01 | 1.95E+01 | Median | 3.44E+01 | 1.78E+01 | 7.93E-01 | 8.49E-02 | 6.28E-01 | 1.96E+01 |
| STD | 2.05E+00 | 1.08E+00 | 2.56E-02 | 7.41E-03 | 3.98E-02 | 9.44E-01 | STD | 1.90E+00 | 8.33E-01 | 2.46E-02 | 5.86E-03 | 4.07E-02 | 6.10E-01 |
| Stacking(ExtraTree+LGBM+RF+KNN+XGB/KNN) | | | | | | | Stacking(ExtraTree+LGBM+RF+KNN+XGB/linear) | | | | | | |
| Metric | RMSE | MAE | R value | MSLE | EVS | SMAPE | Metric | RMSE | MAE | R value | MSLE | EVS | SMAPE |
| Min | 3.61E+01 | 1.89E+01 | 6.68E-01 | 1.05E-01 | 3.86E-01 | 2.19E+01 | Min | 3.19E+01 | 1.69E+01 | 7.23E-01 | 7.80E-02 | 5.16E-01 | 1.89E+01 |
| Max | 4.32E+01 | 2.22E+01 | 7.56E-01 | 1.27E-01 | 5.48E-01 | 2.37E+01 | Max | 4.19E+01 | 2.17E+01 | 8.12E-01 | 1.11E-01 | 6.58E-01 | 2.21E+01 |
| Mean | 4.00E+01 | 2.11E+01 | 7.20E-01 | 1.16E-01 | 4.85E-01 | 2.29E+01 | Mean | 3.53E+01 | 1.88E+01 | 7.77E-01 | 8.99E-02 | 6.00E-01 | 2.04E+01 |
| Median | 3.98E+01 | 2.11E+01 | 7.26E-01 | 1.15E-01 | 4.94E-01 | 2.30E+01 | Median | 3.47E+01 | 1.85E+01 | 7.82E-01 | 8.82E-02 | 6.07E-01 | 2.03E+01 |
| STD | 1.84E+00 | 8.30E-01 | 2.38E-02 | 6.98E-03 | 4.41E-02 | 5.32E-01 | STD | 2.55E+00 | 1.15E+00 | 2.57E-02 | 8.58E-03 | 4.11E-02 | 7.80E-01 |
| Stacking(ExtraTree+LGBM+RF+KNN+XGB/RF) | | | | | | | Stacking(ExtraTree+LGBM+RF+KNN+XGB/SVM) | | | | | | |
| Metric | RMSE | MAE | R value | MSLE | EVS | SMAPE | Metric | RMSE | MAE | R value | MSLE | EVS | SMAPE |
| Min | 3.17E+01 | 1.73E+01 | 7.40E-01 | 7.83E-02 | 5.28E-01 | 1.96E+01 | Min | 3.12E+01 | 1.59E+01 | 7.56E-01 | 7.18E-02 | 5.58E-01 | 1.80E+01 |
| Max | 3.83E+01 | 2.03E+01 | 8.14E-01 | 1.05E-01 | 6.60E-01 | 2.19E+01 | Max | 3.85E+01 | 1.92E+01 | 8.34E-01 | 9.02E-02 | 6.74E-01 | 2.03E+01 |
| Mean | 3.50E+01 | 1.88E+01 | 7.83E-01 | 9.23E-02 | 6.08E-01 | 2.06E+01 | Mean | 3.44E+01 | 1.72E+01 | 8.00E-01 | 7.97E-02 | 6.30E-01 | 1.90E+01 |
| Median | 3.48E+01 | 1.87E+01 | 7.81E-01 | 9.32E-02 | 6.07E-01 | 2.07E+01 | Median | 3.43E+01 | 1.71E+01 | 8.04E-01 | 7.98E-02 | 6.33E-01 | 1.90E+01 |
| STD | 1.85E+00 | 8.25E-01 | 1.97E-02 | 6.92E-03 | 3.45E-02 | 6.10E-01 | STD | 1.93E+00 | 8.29E-01 | 1.74E-02 | 5.79E-03 | 2.69E-02 | 6.61E-01 |
| Stacking(ExtraTree+LGBM+RF+KNN+XGB/XGB) | | | | | | | Stacking(ExtraTree+LGBM+RF+KNN+XGB/ExtraTree) | | | | | | |
| Metric | RMSE | MAE | R value | MSLE | EVS | SMAPE | Metric | RMSE | MAE | R value | MSLE | EVS | SMAPE |
| Min | 3.02E+01 | 1.67E+01 | 7.15E-01 | 7.36E-02 | 5.05E-01 | 1.92E+01 | Min | 3.17E+01 | 1.71E+01 | 7.41E-01 | 7.95E-02 | 5.30E-01 | 1.96E+01 |
| Max | 4.16E+01 | 2.11E+01 | 8.44E-01 | 1.07E-01 | 7.10E-01 | 2.19E+01 | Max | 3.79E+01 | 2.07E+01 | 8.21E-01 | 1.10E-01 | 6.73E-01 | 2.26E+01 |
| Mean | 3.54E+01 | 1.88E+01 | 7.85E-01 | 8.99E-02 | 6.13E-01 | 2.05E+01 | Mean | 3.48E+01 | 1.87E+01 | 7.83E-01 | 9.12E-02 | 6.09E-01 | 2.07E+01 |
| Median | 3.54E+01 | 1.86E+01 | 7.80E-01 | 9.01E-02 | 6.07E-01 | 2.05E+01 | Median | 3.52E+01 | 1.86E+01 | 7.81E-01 | 9.07E-02 | 6.06E-01 | 2.06E+01 |
| STD | 2.43E+00 | 9.42E-01 | 2.79E-02 | 7.63E-03 | 4.57E-02 | 6.40E-01 | STD | 1.98E+00 | 1.00E+00 | 2.08E-02 | 7.84E-03 | 3.44E-02 | 7.71E-01 |

To evaluate the contribution of each component within the best-performing stacking model (ST3), a series of ablation experiments were conducted by incrementally excluding and including individual learners. The prediction accuracy and corresponding MAE for each configuration are illustrated in Figure S2. Initially, the stacking model was tested using only KNN as the base learner, achieving an average R-value of 0.76. When Random Forest (RF) was incorporated into the ensemble, the model's accuracy improved by 2.63%, indicating its significant complementary effect. Further enhancement was observed upon adding LightGBM (LGBM), resulting in an additional 2.56% increase in accuracy. Finally, the inclusion of ExtraTree yielded a substantial improvement of 5.00%, confirming its valuable contribution to the ensemble. These results collectively highlight the additive performance gains achieved through a carefully structured stacking approach.

*Bagging ensemble models finding.* In the second prediction scenario, we developed eight bagging ensemble models selected from 15 ML models based on their individual prediction accuracy. As summarised in Table 6, Bagging Extra-Trees outperformed all other bagging models, achieving an average accuracy of 82.1%, representing a 9% improvement over the standalone Extra-Tree base model. The high performance of Bagging Extra-Trees can be attributed to their randomised splitting mechanism, which enhances generalisation and reduces the risk of overfitting. In contrast, models like XGBoost, CatBoost, and GBR are more susceptible to overfitting, particularly on noisy or imbalanced datasets, unless carefully regularised.

Table 6: Statistical analysis results of the appliances power consumption prediction using four proposed neuro-evolutionary methods.

| | Bag-XGB | | | | | | | Bag-CATB | | | | | |
|---|---|---|---|---|---|---|---|---|---|---|---|---|---|
| Metric | RMSE | MAE | R value | MSLE | EVS | SMAPE | Metric | RMSE | MAE | R value | MSLE | EVS | SMAPE |
| Min | 2.96E+01 | 1.57E+01 | 7.42E-01 | 6.29E-02 | 5.47E-01 | 1.78E+01 | Min | 3.35E+01 | 1.88E+01 | 7.25E-01 | 9.40E-02 | 5.20E-01 | 2.18E+01 |
| Max | 3.85E+01 | 1.96E+01 | 8.66E-01 | 9.12E-02 | 7.39E-01 | 1.99E+01 | Max | 4.16E+01 | 2.25E+01 | 7.95E-01 | 1.20E-01 | 6.21E-01 | 2.43E+01 |
| Mean | 3.31E+01 | 1.71E+01 | 8.09E-01 | 7.73E-02 | 6.51E-01 | 1.90E+01 | Mean | 3.80E+01 | 2.08E+01 | 7.53E-01 | 1.08E-01 | 5.56E-01 | 2.32E+01 |
| Median | 3.24E+01 | 1.69E+01 | 8.13E-01 | 7.70E-02 | 6.56E-01 | 1.90E+01 | Median | 3.82E+01 | 2.09E+01 | 7.49E-01 | 1.08E-01 | 5.46E-01 | 2.32E+01 |
| STD | 2.68E+00 | 1.05E+00 | 2.87E-02 | 7.07E-03 | 4.43E-02 | 6.03E-01 | STD | 2.23E+00 | 9.13E-01 | 2.34E-02 | 6.09E-03 | 3.26E-02 | 6.44E-01 |
| | Bag-DT | | | | | | | Bag-ExtraTree | | | | | |
| Metric | RMSE | MAE | R value | MSLE | EVS | SMAPE | Metric | RMSE | MAE | R value | MSLE | EVS | SMAPE |
| Min | 2.85E+01 | 1.56E+01 | 7.69E-01 | 6.50E-02 | 5.90E-01 | 1.76E+01 | Min | 2.96E+01 | 1.62E+01 | 7.85E-01 | 6.70E-02 | 6.14E-01 | 1.81E+01 |
| Max | 3.72E+01 | 1.92E+01 | 8.48E-01 | 9.21E-02 | 7.17E-01 | 2.06E+01 | Max | 3.72E+01 | 1.81E+01 | 8.56E-01 | 8.71E-02 | 7.30E-01 | 1.96E+01 |
| Mean | 3.26E+01 | 1.69E+01 | 8.09E-01 | 7.69E-02 | 6.52E-01 | 1.89E+01 | Mean | 3.28E+01 | 1.70E+01 | **8.21E-01** | 7.66E-02 | 6.72E-01 | 1.88E+01 |
| Median | 3.29E+01 | 1.70E+01 | 8.11E-01 | 7.70E-02 | 6.56E-01 | 1.91E+01 | Median | 3.28E+01 | 1.71E+01 | 8.22E-01 | 7.71E-02 | 6.75E-01 | 1.88E+01 |
| STD | 2.39E+00 | 9.73E-01 | 2.39E-02 | 6.82E-03 | 3.77E-02 | 6.87E-01 | STD | 2.04E+00 | 5.64E-01 | 1.95E-02 | 5.40E-03 | 3.16E-02 | 4.58E-01 |
| | Bag-GBR | | | | | | | Bag-LGBM | | | | | |
| Metric | RMSE | MAE | R value | MSLE | EVS | SMAPE | Metric | RMSE | MAE | R value | MSLE | EVS | SMAPE |
| Min | 3.86E+01 | 2.30E+01 | 5.96E-01 | 1.24E-01 | 3.50E-01 | 2.58E+01 | Min | 3.23E+01 | 1.91E+01 | 6.85E-01 | 9.71E-02 | 4.67E-01 | 2.21E+01 |
| Max | 4.73E+01 | 2.66E+01 | 7.19E-01 | 1.59E-01 | 4.94E-01 | 2.85E+01 | Max | 4.43E+01 | 2.38E+01 | 7.85E-01 | 1.30E-01 | 6.08E-01 | 2.55E+01 |
| Mean | 4.27E+01 | 2.46E+01 | 6.61E-01 | 1.43E-01 | 4.27E-01 | 2.73E+01 | Mean | 3.83E+01 | 2.14E+01 | 7.36E-01 | 1.14E-01 | 5.35E-01 | 2.38E+01 |
| Median | 4.31E+01 | 2.45E+01 | 6.60E-01 | 1.43E-01 | 4.27E-01 | 2.73E+01 | Median | 3.83E+01 | 2.14E+01 | 7.43E-01 | 1.14E-01 | 5.41E-01 | 2.39E+01 |
| STD | 2.60E+00 | 1.00E+00 | 2.45E-02 | 9.92E-03 | 2.88E-02 | 6.90E-01 | STD | 2.93E+00 | 1.25E+00 | 2.55E-02 | 8.43E-03 | 3.54E-02 | 7.63E-01 |
| | Bag-RF | | | | | | | Bag-KNN | | | | | |
| Metric | RMSE | MAE | R value | MSLE | EVS | SMAPE | Metric | RMSE | MAE | R value | MSLE | EVS | SMAPE |
| Min | 3.01E+01 | 1.65E+01 | 7.67E-01 | 6.98E-02 | 5.80E-01 | 1.88E+01 | Min | 2.98E+01 | 1.55E+01 | 7.90E-01 | 7.01E-02 | 6.21E-01 | 1.76E+01 |
| Max | 3.69E+01 | 1.91E+01 | 8.33E-01 | 9.61E-02 | 6.71E-01 | 2.13E+01 | Max | 3.74E+01 | 1.79E+01 | 8.35E-01 | 8.62E-02 | 6.89E-01 | 1.91E+01 |
| Mean | 3.36E+01 | 1.79E+01 | 7.99E-01 | 8.38E-02 | 6.29E-01 | 2.01E+01 | Mean | 3.26E+01 | 1.66E+01 | 8.08E-01 | 7.58E-02 | 6.52E-01 | 1.85E+01 |
| Median | 3.39E+01 | 1.80E+01 | 8.01E-01 | 8.50E-02 | 6.30E-01 | 2.02E+01 | Median | 3.26E+01 | 1.65E+01 | 8.08E-01 | 7.48E-02 | 6.51E-01 | 1.85E+01 |
| STD | 1.94E+00 | 7.82E-01 | 1.80E-02 | 7.32E-03 | 2.57E-02 | 6.90E-01 | STD | 1.99E+00 | 6.66E-01 | 1.32E-02 | 4.79E-03 | 2.02E-02 | 4.72E-01 |

To assess the effect of the number of estimators on the performance of bagging ensembles, we conducted a detailed experiment using Bagging with Extra Trees (Bag-ExtraTree, best-performed) and XGBoost (BagXGB) as base learners. Each model was evaluated across a range of ensemble sizes, varying the number of estimators from 1 to 30. As can be illustrated in Figure S1, increasing the number of estimators initially leads to improvements in both prediction accuracy (R-value) and MAE, indicating enhanced generalisation and reduced prediction error. However, this trend does not persist until 30. In the case of Bag-ExtraTree, performance gains plateau after 14 estimators, while Bag-XGB shows diminishing returns beyond 24 estimators. These observations highlight the importance of selecting an optimal ensemble size to avoid unnecessary computational complexity without compromising model accuracy.

*Voting ensemble models finding.* Table 7 presents the statistical prediction results of six voting ensemble models. Among these, the combination of Extra-Trees and LGBM in a bagging framework achieved the highest average accuracy of 80.6%. This superior performance can be attributed to the complementary strengths of the two algorithms, as their diversity and aggregation enhance overall predictive capabilities.

The box-and-whisker plot in

Table 7: Statistical analysis results of the appliances power consumption prediction using six proposed voting ensemble methods.

| | Voting (XGB+LGBM) | | | | | | Voting (XGB+CATB) | | | | | |
|---|---|---|---|---|---|---|---|---|---|---|---|---|
| Metric | RMSE | MAE | R value | MSLE | EVS | SMAPE | Metric | RMSE | MAE | R value | MSLE | EVS | SMAPE |
| Min | 3.23E+01 | 1.73E+01 | 7.69E-01 | 7.61E-02 | 5.87E-01 | 1.92E+01 | Min | 2.86E+01 | 1.69E+01 | 7.04E-01 | 7.74E-02 | 4.94E-01 | 1.93E+01 |
| Max | 3.68E+01 | 1.97E+01 | 8.42E-01 | 9.46E-02 | 7.06E-01 | 2.10E+01 | Max | 3.84E+01 | 2.04E+01 | 8.24E-01 | 1.02E-01 | 6.74E-01 | 2.18E+01 |
| Mean | 3.42E+01 | 1.82E+01 | 7.92E-01 | 8.41E-02 | 6.27E-01 | 2.00E+01 | Mean | 3.47E+01 | 1.85E+01 | 7.85E-01 | 8.79E-02 | 6.14E-01 | 2.05E+01 |
| Median | 3.40E+01 | 1.81E+01 | 7.89E-01 | 8.49E-02 | 6.22E-01 | 1.99E+01 | Median | 3.56E+01 | 1.83E+01 | 7.87E-01 | 8.67E-02 | 6.17E-01 | 2.04E+01 |
| STD | 1.43E+00 | 6.35E-01 | 2.05E-02 | 4.70E-03 | 3.24E-02 | 4.71E-01 | STD | 2.64E+00 | 1.14E+00 | 2.60E-02 | 8.02E-03 | 3.94E-02 | 6.92E-01 |
| | Voting (XGB+KNN) | | | | | | Voting (ExtraTree+LGBM) | | | | | |
| Metric | RMSE | MAE | R value | MSLE | EVS | SMAPE | Metric | RMSE | MAE | R value | MSLE | EVS | SMAPE |
| Min | 3.02E+01 | 1.60E+01 | 7.38E-01 | 7.17E-02 | 5.36E-01 | 1.84E+01 | Min | 2.91E+01 | 1.61E+01 | 7.65E-01 | 7.25E-02 | 5.85E-01 | 1.85E+01 |
| Max | 3.85E+01 | 1.94E+01 | 8.40E-01 | 9.86E-02 | 7.01E-01 | 2.10E+01 | Max | 3.67E+01 | 1.91E+01 | 8.48E-01 | 1.02E-01 | 7.17E-01 | 2.11E+01 |
| Mean | 3.43E+01 | 1.75E+01 | 7.98E-01 | 8.28E-02 | 6.35E-01 | 1.93E+01 | Mean | 3.34E+01 | 1.75E+01 | **8.06E-01** | 8.32E-02 | 6.48E-01 | 1.96E+01 |
| Median | 3.40E+01 | 1.73E+01 | 7.95E-01 | 8.27E-02 | 6.32E-01 | 1.91E+01 | Median | 3.31E+01 | 1.74E+01 | 8.05E-01 | 8.08E-02 | 6.43E-01 | 1.94E+01 |
| STD | 2.19E+00 | 8.50E-01 | 2.43E-02 | 7.51E-03 | 3.93E-02 | 6.37E-01 | STD | 2.10E+00 | 9.03E-01 | 2.15E-02 | 8.19E-03 | 3.44E-02 | 7.93E-01 |
| | Voting (ExtraTree+CATB) | | | | | | Voting (ExtraTree+KNN) | | | | | |
| Metric | RMSE | MAE | R value | MSLE | EVS | SMAPE | Metric | RMSE | MAE | R value | MSLE | EVS | SMAPE |
| Min | 2.99E+01 | 1.64E+01 | 7.63E-01 | 7.01E-02 | 5.79E-01 | 1.86E+01 | Min | 2.87E+01 | 1.54E+01 | 7.51E-01 | 6.66E-02 | 5.58E-01 | 1.79E+01 |
| Max | 3.72E+01 | 1.94E+01 | 8.23E-01 | 9.29E-02 | 6.76E-01 | 2.09E+01 | Max | 3.84E+01 | 1.93E+01 | 8.48E-01 | 9.69E-02 | 7.19E-01 | 2.03E+01 |
| Mean | 3.39E+01 | 1.80E+01 | 7.98E-01 | 8.35E-02 | 6.35E-01 | 1.99E+01 | Mean | 3.40E+01 | 1.74E+01 | 7.98E-01 | 8.27E-02 | 6.34E-01 | 1.90E+01 |
| Median | 3.42E+01 | 1.82E+01 | 8.03E-01 | 8.38E-02 | 6.41E-01 | 2.00E+01 | Median | 3.45E+01 | 1.77E+01 | 7.96E-01 | 8.31E-02 | 6.33E-01 | 1.89E+01 |
| STD | 1.98E+00 | 7.98E-01 | 1.83E-02 | 6.06E-03 | 2.89E-02 | 5.58E-01 | STD | 2.89E+00 | 1.21E+00 | 2.31E-02 | 9.19E-03 | 3.82E-02 | 7.82E-01 |

*4.3. Visual Interpretation and Model Performance Insights*

This section provides a qualitative look at the most significant experimental results to accompany the quantitative findings in the previous section. Using various plots, model behaviour comparisons, and performance visualizations, we aspire to provide deeper insight into the predictive ability and interpretability of the proposed ensemble learning models. The visualisations help identify temporal trends, model robustness, and relative performance of different configurations under actual real-world smart building conditions.

*4.3.1. Benchmark Models Results*

The box-and-whisker plot 9 presented in this analysis offers a comprehensive evaluation of 14 machine and deep learning techniques utilised for predicting household appliance energy consumption. This evaluation focuses on prediction accuracy and MAE. In plot 9, a box is drawn between the first and third quartiles, with a vertical line passing through the box at the median. The whiskers extend from each quartile to the minimum and maximum values. In addition, any outliers in the dataset are depicted as single red crosses on the diagram. Upon analysing the plot, it becomes evident that XGBoost consistently outperforms other models in terms of the median accuracy metric. Following XGBoost, the GBM and Catboost models exhibit comparable performances. Furthermore, an intriguing observation can be made regarding the effectiveness of adding a convolutional layer to a dense model. This enhancement significantly improves the average performance of the model, indicating its potential for achieving higher accuracy. These findings provide valuable insights into the comparative performance of different machine and deep learning techniques in predicting household appliance energy consumption. This information can aid researchers and practitioners in selecting the most suitable models for their specific purposes, thereby enhancing the accuracy of energy consumption predictions. Furthermore, Figure 9(b) presents the average absolute validation error for a set of 14 ML models. In terms of MSE, XGBoost stands out as the top performer with an impressive score of 32. Notably, the EBM model showcases a competitive performance in MAE and secures the second rank. Moreover, the GBM, Random Forest, and CDNN models also demonstrate noteworthy performances, yielding acceptable results in their respective evaluations. This information provides valuable insights and highlights the strengths of XGBoost in achieving low MSE while acknowledging the competitive performance of the EBM model in MAE. The notable performances of GBM, Random Forest, and CDNN further contribute to the range of acceptable results obtained. These findings assist in understanding the efficacy of different ML models and offer guidance for selecting the most suitable approach based on the desired evaluation metric.

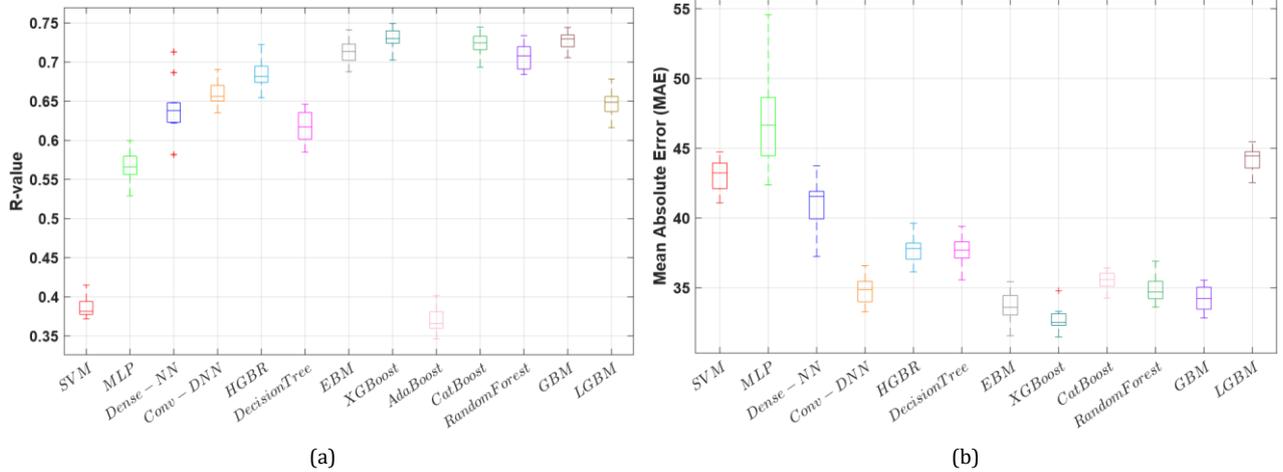

Figure 9: The box-and-whisker plot of statistical results evaluation for 14 machine and deep learning techniques used for predicting the energy consumption of household appliances, based on a) prediction accuracy (R-value) and b) mean absolute error (MAE).

*4.3.2. Ensemble Models Findings*

Figure 10 presents the statistical performance of 10 stacking models (listed in Table S4) evaluated in terms of R-value and MAE. Among these, the best-performing model in terms of median R-value accuracy is ST-M3 (ExtraTree+LGBM+RF+KNN/MLP), achieving an accuracy of 81%. Conversely, the model STM8 (ExtraTree+LGBM+RF+KNN+XGB/SVM) exhibits the lowest median MAE, accurately predicting appliance power consumption with a value of approximately 17.1.

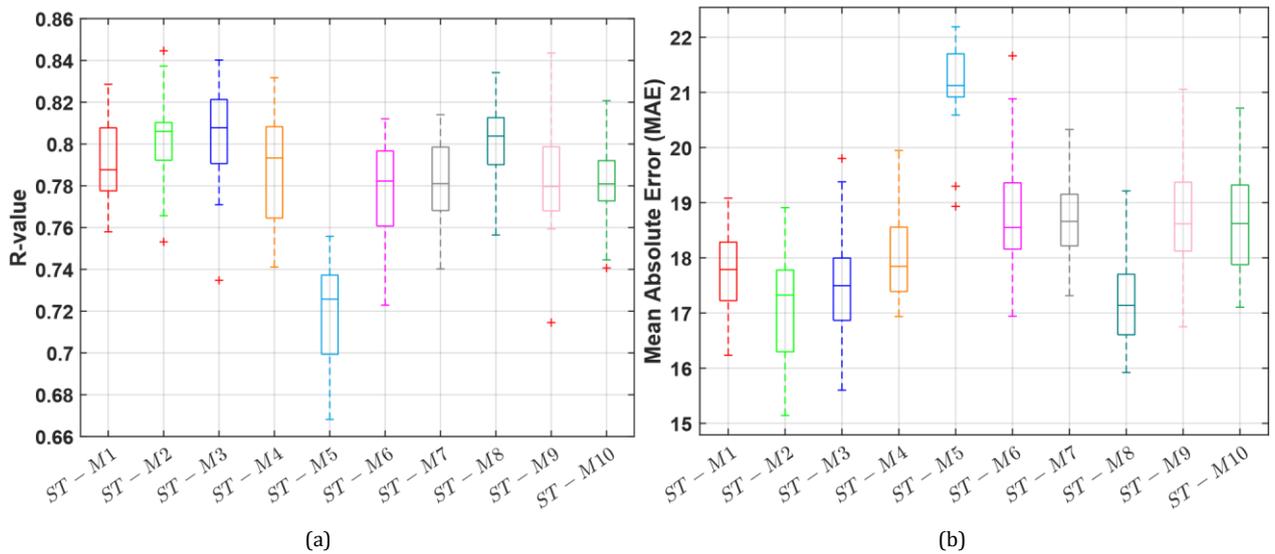

Figure 10: The box-and-whisker plot of (a) R-value and b) MAE statistical results for the ten stacking ensemble method in predicting the energy consumption of appliances in the smart house.

Figure 11 provides a detailed comparison of the eight bagging models in terms of R-value and MAE. While Bagging KNN demonstrated the lowest average MAE among all models, its overall accuracy was lower than that of Bagging Extra-Trees, Decision Trees, and XGBoost. This highlights a trade-off between minimising error and maximising accuracy, with Bagging Extra-Trees striking the best balance among the evaluated models.

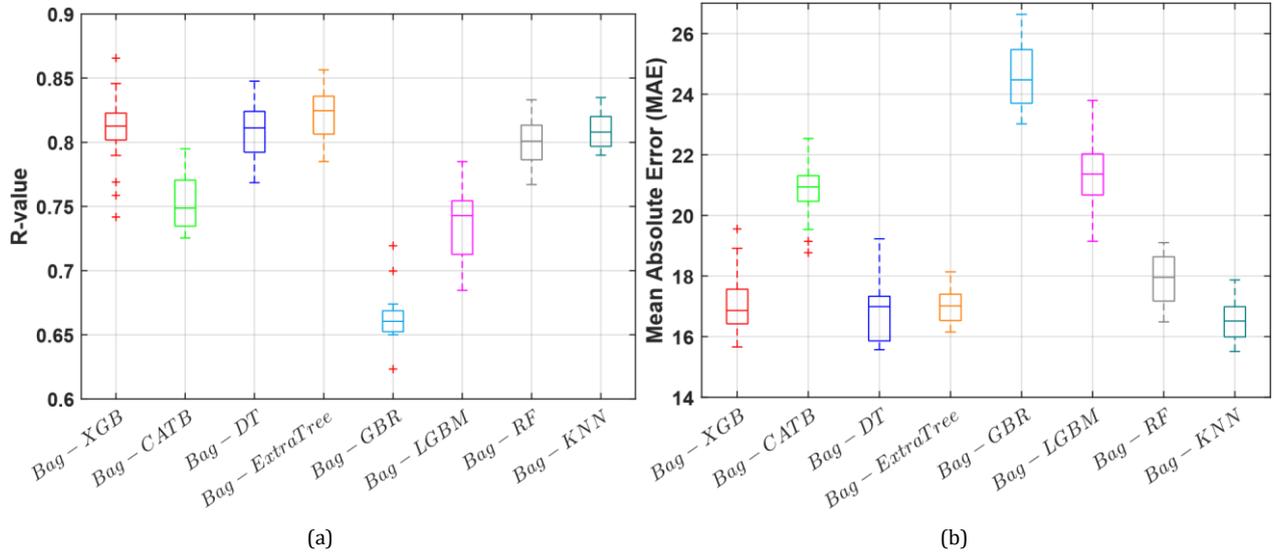

(a)          (b)

Figure 11: The box-and-whisker plot of (a) R-value and b) MAE statistical results for eight bagging ensemble method in predicting the energy consumption of appliances in the smart house.

Figure 12 illustrates the performance of these models in terms of R-value and MAE. While Voting(XGB+LGBM) achieved the best median R-value, Voting(Extra-Tree+KNN) outperformed other models with the lowest average MAE, demonstrating its effectiveness in minimising prediction error.

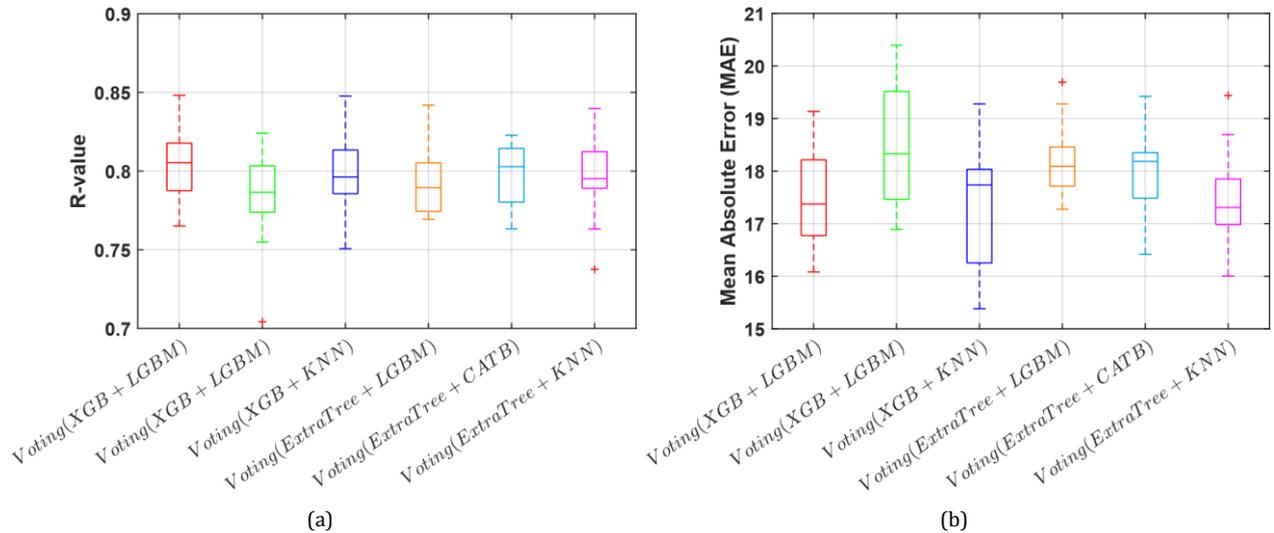

(a)          (b)

Figure 12: The box-and-whisker plot of (a) R-value and b) MAE statistical results for six voting ensemble methods in predicting the energy consumption of appliances in the smart house.

*4.3.3. Final Comparisons*

To ensure a fair comparison among the ensemble models proposed in this study, the results are presented in Figure 13. As observed, the Bagging Extra-Trees model significantly outperformed the other ensemble methods, with a p-value less than 0.05 for both accuracy and MAE, indicating its superior predictive performance. Bagging ensembles are particularly effective in scenarios requiring variance reduction, noise handling, and robust generalisation across diverse datasets. These characteristics make bagging an ideal choice for predicting appliance power consumption, outperforming voting and stacking models in this context.

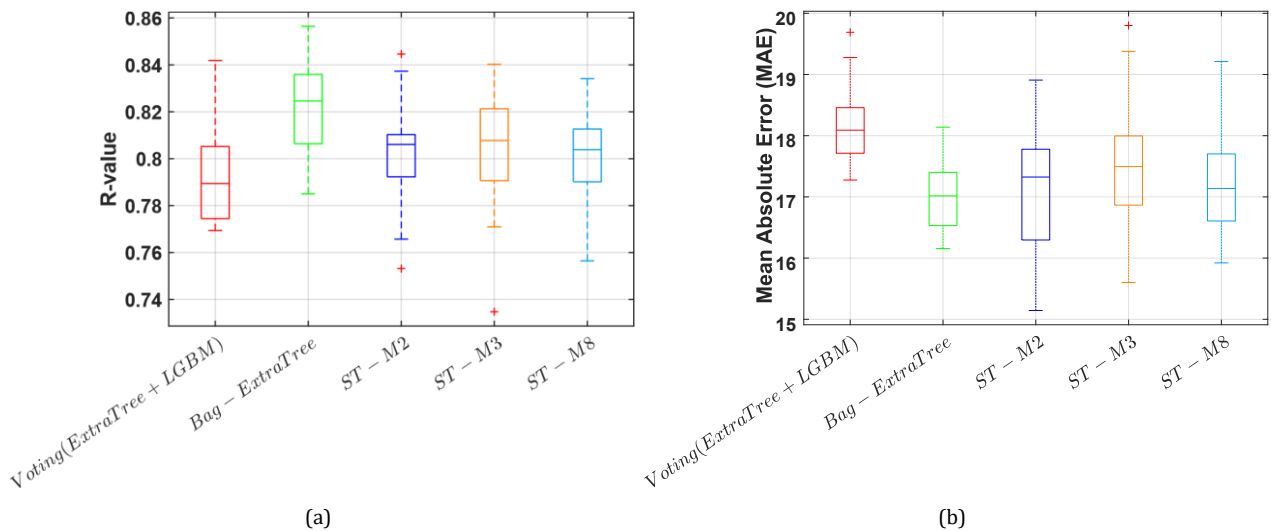

(a)          (b)

Figure 13: The box-and-whisker plot of (a) R-value and b) MAE statistical results for best-performed voting, bagging and stacking ensemble methods in predicting the energy consumption of appliances in the smart house.

The results of the experiment demonstrate the superiority of the ExtraTree Bagging ensemble model over the Stacking (ST-M2, ST-M3, and ST-M8) and Voting ensemble methods. Specifically, the Bagging model achieved the best prediction accuracy rates for all the performance metrics. This is because the nature of Bagging reduces variance by combining the predictions of several decorrelated ExtraTree base models trained on different bootstrap samples. The ExtraTrees' randomness encourages model diversity and generalisation, thus, more stable and precise predictions. The Stacking model, however, relies on a metamodel to combine base models, and this sometimes can introduce additional bias and can be susceptible to overfitting if not carefully tuned. The Voting ensemble, similarly, treats all base learners equally without dynamically leveraging their individual strengths. These findings confirm that Bagging architecture, coupled with ExtraTrees provides a more robust and stable solution for energy consumption prediction in smart buildings.

*4.3.4. Comparison with Other Techniques*

To ensure a comprehensive comparison with previous studies using similar datasets, we evaluated 19 machine learning models adopted from the works of Candanedo et al. [32] and Han et al. [68]. These models include Affinity Propagation Radial Basis Function (AP-RBF), standard Radial Basis Function (RBF) networks, and Backpropagation (BP) neural networks [68], each tested under varying configurations of hidden nodes to enable a robust and consistent performance assessment.

Figure 14 provides a comparative assessment of RMSE scores of a variety of prediction models applied to the same dataset. Among all the models, our proposed model, Voting (XGB+KNN) ensemble produced the lowest RMSE, indicating superior predictive accuracy. It was closely followed by Bag-ExtraTree and ST-M3, both of which also performed well with significantly lower error rates than their standard base models. On the other hand, models such as AP-BP [68] and AP-ELM [68] possessed the highest RMSE values, which signifies poor generalisation ability and fitness to the target data. Ensemble methods such as XGB, CatB, and HGBR performed better than individual models such as SVM-Radial [32], GBM [32], and RF [32] time and again, reinforcing the advantage of ensemble creation for increasing the predictability robustness.

Besides, neural-based architectures like MLP, DNN, and BP of reasonable sizes acted competitively but were sensitive to network size and training dynamics. Overall, the results show that ensemble and hybrid strategies are highly effective in controlling prediction errors in this application field.

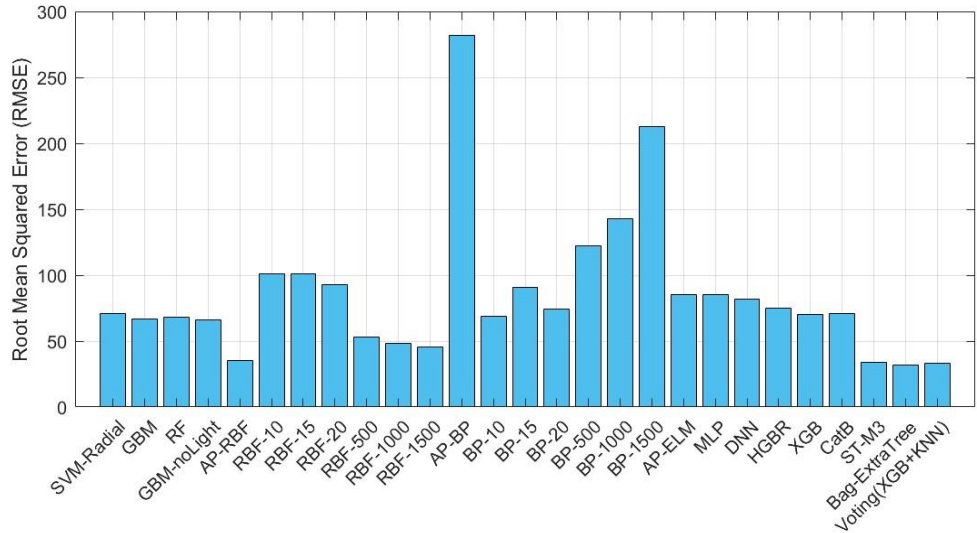

Figure 14: Comparative analysis of energy consumption forecasting between the proposed models and prior studies.

*4.4. Hyper-parameters optimisation*

Hyper-parameters play a critical role in enhancing the performance of prediction methods, as they govern key aspects of the model's learning process, such as complexity, learning rate, regularization strength, and optimization strategy. Proper tuning of these parameters can significantly improve a model's accuracy [69]. To evaluate the impact of hyper-parameters on model performance, we conducted an analysis using a greedy search, focusing on four key hyper-parameters: the number of estimators for Extra-Trees and Bagging, along with the maximum rate of features and samples used during training. The optimisation landscape for the number of estimators in Extra-Trees and Bagging is depicted in Figure 15(a). For the Bagging ensemble, the number of estimators was evaluated in the range of 5 to 50, while for the Extra Trees model, the range of 10 to 100 was tested. The highest prediction accuracy was achieved with Bagging at Ns = 15, and with Extra Trees when the number of estimators exceeded 60. The results indicate that the number of estimators in the Bagging model has a more substantial influence on achieving higher accuracy compared to the number of estimators in the Extra-Trees model. Figure 15(b) illustrates the prediction accuracy across different configurations of maximum sample rate and feature rate. The results indicate that the highest accuracy is obtained when both parameters exceed a threshold of 0.6, suggesting that retaining a larger proportion of samples and features enhances model performance. This highlights the critical role of properly tuning the Bagging model's hyper-parameters for improved predictive performance.

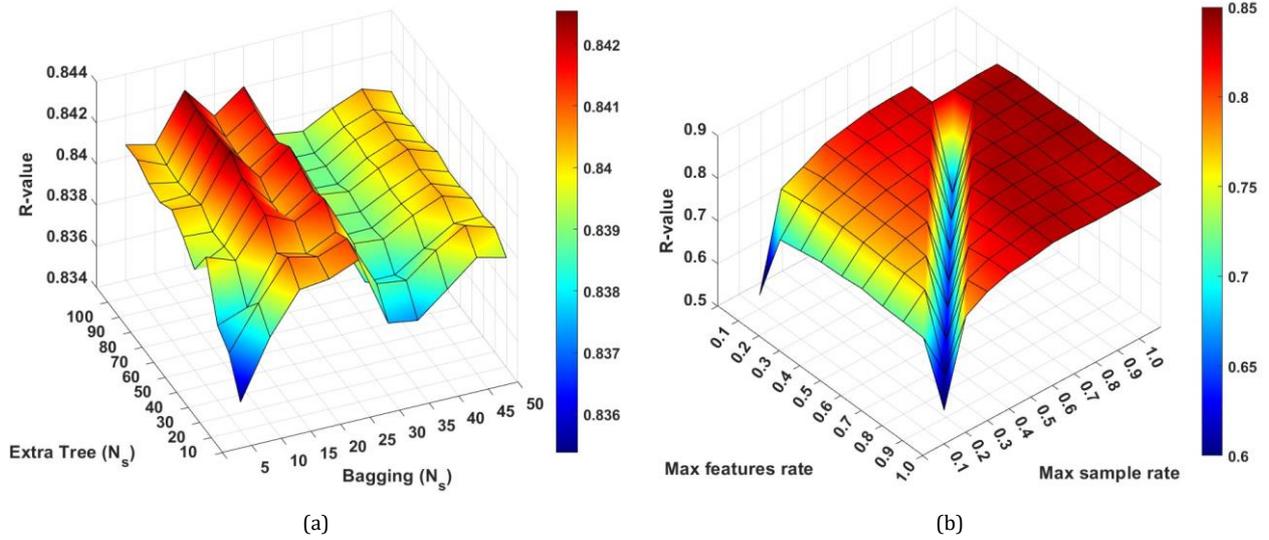

Figure 15: Hyper-parameters tuning using grid search for bagging Extra Tree ensemble model

In this study, we used four effective and well-known optimisation methods in order to adjust the hyperparameters of ensemble models. In the first step, we focused on XGBoost hyper-parameters optimisation and they are listed in Table S2. Figure 16 illustrates a comparison of the average convergence speeds exhibited by these optimisation methods. It is important to note that the population size and maximum evaluation number are consistent across all methods at 25 and 1000, respectively. Upon analysis of Figure 16, it is evident that XGB-EA demonstrates rapid convergence towards a semi-optimal configuration of hyperparameters within the initial 20% of the total evaluation count. However, XGB-EA encounters challenges when confronted with a local optimum, and the mutation strategy employed does not effectively facilitate the exploration of alternative feasible regions. Conversely, although XGB-DE initially displayed a convergence rate lower than that of XGB-PSO and XGB-GA during the exploration phase, it ultimately managed to discover superior solutions. Considering the computational expense and time consumption associated with training the model, we recommend employing the 1+1EA meta-heuristic as a hyper-parameter optimiser.

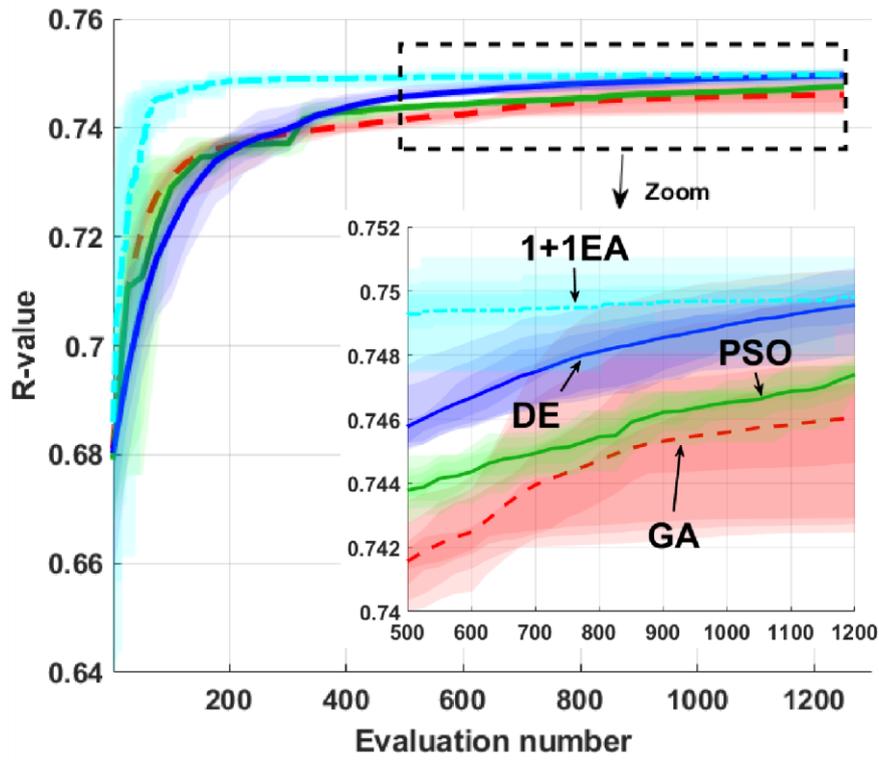

(a)

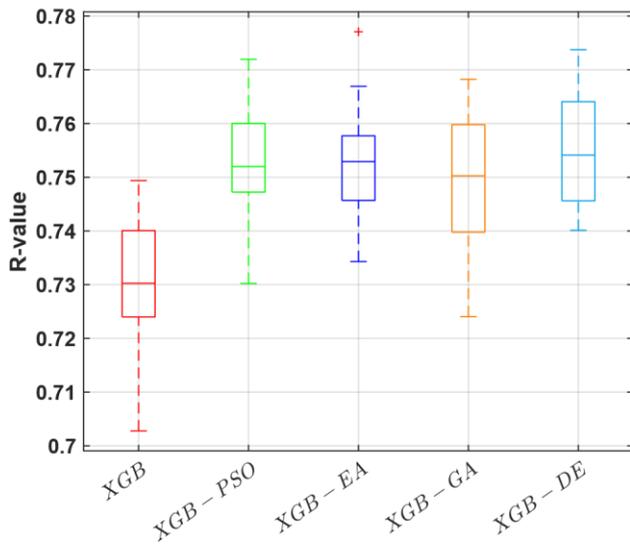

(b)

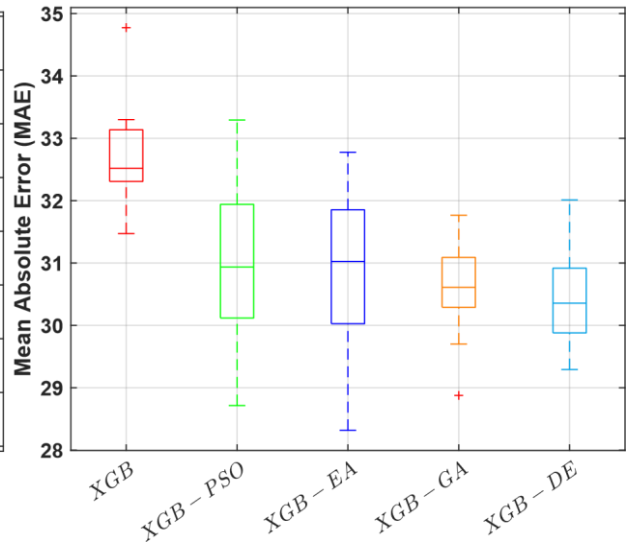

(c)

Figure 16: a) A convergence rate comparison for four neuro-evolutionary algorithms including XGB-GA, XGB-DE, XGB-PSO, and XGB-EA. The lines show the average accuracy achieved by whole solutions in each generation.

Moreover, Figure 16 (b) and (c) shows the statistical performance analysis of the XGBoost with predefined hyper-parameters and four proposed neuro-evolutionary methods in terms of accuracy and MAE. It is crystal clear that the best-performing hybrid model is XGB-DE in terms of metrics, accuracy, and MAE. The accuracy and MAE improvement of XGB-DE are 3.5% and 7.6% compared with XGBoost.

Table S5 reports more technical comparison results of four evolutionary ensemble models. We can see that XGB-DE prediction results had the minimum distance with the true power consumption values confirmed by metrics RMSE, MAE, and MSLE. In terms of correlation coefficient (R-value), all hybrid models competitively performed well; however, XGB-DE outperformed other models. Finally, we evaluated the performance of four optimisation methods to enhance the Bagging Extra-Trees , best-performed model, as shown in Figure 17. Among the tested methods, 1+1EA (Bag-ET-EA) demonstrated the fastest convergence during the initial iterations, highlighting its efficiency in optimisation. This experiment confirms that 1+1EA is an effective optimiser for fine-tuning hyper-parameters. Additionally, the balance between exploration and exploitation for the four hyper-parameters is illustrated in Figures 17(b-e), providing further insights into the optimisation dynamics of each method.

As can be seen from Figure 17(b), the optimisation process commenced by exploring a wide range of values for the number of Bagging estimators, ranging from 10 to 90. Throughout successive iterations, this search space increasingly narrowed, echoing the transition from exploration to exploitation, and finally converged within an optimum range between 60 and 65. An identical convergence pattern could be observed for the maximum feature rate hyperparameter, plotted in Figure 17(c), where the search process converged around the value of 0.4. At the highest sample rate (Figure 17(d), the optimiser found good performing regions early in the search and converged rapidly to values above 0.9, finally settling at 1. Moreover, the number of estimations was subjected to an extensive and dense search over a larger space, with over 200 evaluations. Despite the wide initial range, the optimiser focused on configurations from above 60 estimators onwards and eventually settled at 80. Results like these bear testament to the optimiser's fair balance of local refinement and global search, terminating at well-chosen hyper-parameters for better model performance.

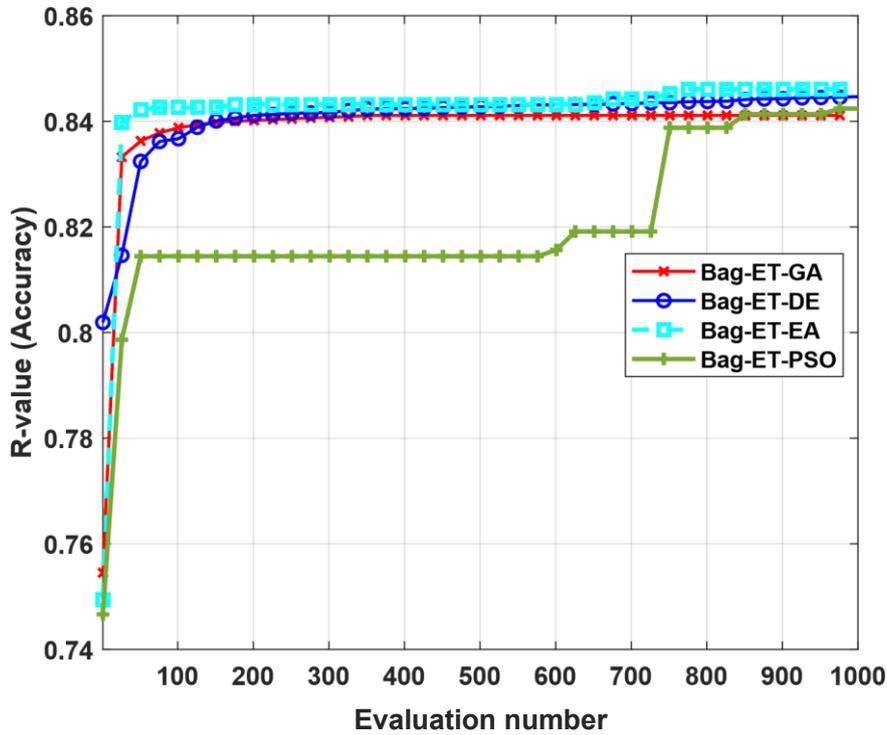

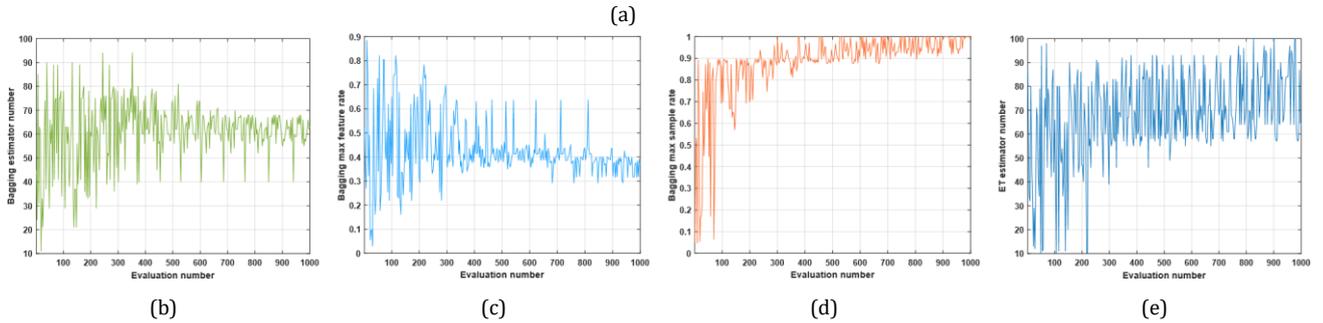

Figure 17: a) Convergence rate of Bagging Extra-tree's hyper-parameters tuning using four optimisation methods and the exploration of parameters search space b) Bagging estimator number, c) maximum feature rate of Bagging, d) maximum sample rate of Bagging, and e) estimator number of Extra tree method.

## 5. Discussions and Future directions

The proposed hybrid evolutionary ensemble models offer principal advantages in predicting total power consumption in smart buildings by effectively harnessing the merits of diverse learning algorithms and strong evolutionary optimisation. By integrating ensemble techniques such as Bagging, Stacking, and Voting with adaptive metaheuristic-based hyper-parameter tuning, the models achieve better accuracy, stability, and generalisations on highly dynamic and nonlinear energy consumption patterns. The hybrid approach enables the model to capture sophisticated dependencies between weather conditions, occupancy patterns, appliance use, and ambient factors, typically neglected by separate algorithms. Moreover, the evolutionary optimisation process intelligence searches the hyper-parameter space, free from hand-tuning, and circumvents possible overfitting.

*5.1. Scalability and Dynamic pattern*

The adaptive ensemble evolution learning method demonstrated in the proposal holds high scalability potential for use across various smart building environments with varying occupancy behaviour and energy use patterns. This is due to the modularity of the model, where multiple base learners (ExtraTrees, XGBoost, LGBM) are blended across ensemble frameworks (Bagging, Stacking, and Voting) and leverage evolutionary algorithms to drive optimisation of the hyperparameters. The combination of diverse learning paradigms enables the model to learn linear and nonlinear energy consumption patterns, and the evolutionary optimisation adjusts hyper-parameters according to different building-specific data distributions. These capabilities put the model in a position to generalise well beyond the current test case, particularly when retrained on new data from buildings with different spatial configurations, climate regions, or operating schedules.

Furthermore, the hybrid dataset used in this research, indoor and outdoor temperature, humidity, lighting, occupancy, and appliance-level usage, represents a realistic and comprehensive sensing environment that is becoming more common in modern smart buildings. The evolutionary tuning process also enables the model to adapt dynamically to changes in input feature importance, e.g., peak-hour demand or seasonal trends, which makes it more robust across various environments. Therefore, the model proposed is not limited to the Belgian building that was used to evaluate but can also be generalised to other types of buildings like commercial offices, schools, or housing estates. Follow-up work will focus on testing the generality by means of transfer learning techniques and cross-building training data in order to provide global deployment of the model towards energy prediction and management in varied smart building setups.

*5.2. Real-time and Computational Efficiency*

The proposed adaptive models possess great potential for real-time deployment in smart building environments. By leveraging the use of lightweight learners such as Extra Trees within a Bagging framework and adjusting the parameters using computationally lightweight metaheuristic algorithms such as the 1+1 EA, the computational overhead at both the training and inference phases is reduced significantly. Due to its parallelisable nature, the Bagging framework facilitates simultaneous training and independent operation of numerous base models, making scaling simpler on multi-core or distributed systems. Additionally, the evolutionary optimisation method accelerates convergence to optimal model configurations by efficiently exploring the search space, which decreases the number of training iterations. These qualities make the proposed models highly suitable for real-time or near-real-time energy forecasting, where quick adaptation to new sensor readings is essential for dynamic energy management and demand-side response in smart buildings.

Additionally, the framework's computational efficiency was verified by monitoring training and prediction run times during cross-validation experiments. Compared to traditional ensemble models such as boosting models (e.g., XGBoost, CatBoost, GBM) that involve sequential model updating and longer processing, the proposed Bagging-based model, enhanced by evolutionary tuning, consistently had lower computational costs without sacrificing predictive accuracy. This accuracy-efficiency trade-off ensures the practical viability of deploying the model in real building management systems, where timely forecasting is crucial for energy scheduling, load balancing, and integration with renewable sources. Thus, the hybrid evolutionary ensemble method improves the forecasting accuracy and meets the operational requirements of smart building applications in terms of speed, scalability, and resource efficiency.

*5.3. Future Directions*

Future research will focus on enhancing the applicability and robustness of the proposed adaptive evolutionary ensemble models by their broader implementation in different building typologies and climatic zones. This will be realised by integrating diversified, large-scale datasets with varying occupancy schedules, appliance utilisation profiles, architectural features, types of HVAC systems, and external environmental factors such as solar irradiance, wind speed, and air quality. By including a more extensive set of input features, the model will generalise better to residential, commercial, and institutional buildings with different temporal and spatial patterns of energy consumption. In addition, an effort will be made to integrate real-time data streaming into the prediction pipeline so that the model can be executed in an online learning mode. This will allow the forecasting engine to adjust its parameters in real-time as it receives new sensor data, in order to deliver more accurate and responsive control in dynamic energy management systems.

Advanced optimisation techniques, such as multi-objective evolutionary algorithms, cooperative coevolution, and meta-reinforcement learning, will be explored to attain further improvements in model convergence speed, scalability, and flexibility. Finally, incorporating renewable energy forecasting, such as photovoltaic and wind power generation, into the ensemble framework will help develop smart, carbon-aware decisionmaking systems. These enhancements will not only improve forecast accuracy but also enable real-time load balancing, demand-side management, and ultimately the decarbonisation and sustainability of future smart buildings.

Future research will also focus on applying the model developed to other forms of smart buildings with varying configurations and usage patterns. To enhance the objectivity and generalizability of the model, we also intend to incorporate standardized building classification systems and develop a taxonomy-based modelling process that accounts for variations in room types, appliance densities, and user usage patterns. In addition, applying the framework to multi-building datasets will provide cross-building validation and more scalable and policy-relevant energy forecasting solutions.

## 6. Conclusion

In conclusion, the building sector accounts for a significant portion of global energy consumption and plays a crucial role in future decarbonisation efforts. Therefore, developing reliable and accurate energy demand forecasting models is essential to effectively manage energy consumption and improve energy efficiency in smart buildings.

This paper addresses the challenges of predicting total energy use in smart buildings, complicated by temporal oscillations and complex linear and non-linear patterns. To overcome these challenges, the paper proposes three adaptive evolutionary ensemble models that integrate various bagging, stacking and voting models with a fast and effective evolutionary hyper-parameters tuner. Data filtering and automatic outlier removal techniques were also employed to extract relevant parameters and enhance prediction accuracy.

The proposed energy forecasting model was evaluated using a hybrid dataset encompassing meteorological parameters, appliance energy use, temperature, humidity, and lighting energy consumption data collected from 18 sensors in a Stambruges, Mons, Belgium building. To benchmark the performance of the proposed model, it was compared against 15 popular ML models, including classic ML models, neural networks, decision trees, random forests, deep learning models, and ensemble models. The findings demonstrate that the adaptive evolutionary bagging model outperformed the other prediction models in terms of accuracy and learning error. Specifically, it achieved accuracy improvements of 12.6%, 13.7%, 12.9%, 27.04%, and 17.4% compared to XGB, CatBoost, GBM, LGBM, and RF, respectively. These results highlight the effectiveness of the advanced evolutionary ensemble approach for energy demand forecasting in intelligent buildings. By surpassing the performance of various established ML models, the proposed model showcases its potential to enhance prediction accuracy and contribute to efficient energy management in smart buildings.

**Declaration of competing interest**

The authors declare that they have no known competing financial interests or personal relationships that could have appeared to influence the work reported in this paper.

## Supplementary Material

Table S1: The list and description of features collected from the building

| Location | Ref | Full name | Units | Location | Ref | Full name | Units |
|---|---|---|---|---|---|---|---|
| Kitchen | T1<br>RH1 | Temperature<br>Humidity | ◦C<br>% | Teenager room 2 | T8<br>RH8 | Temperature<br>Humidity | ◦C<br>% |
| Living room | T2<br>RH2 | Temperature<br>Humidity | ◦C<br>% | Parents room | T9<br>RH9 | Temperature<br>Humidity | ◦C<br>% |
| Laundry room | T3<br>RH3 | Temperature<br>Humidity | ◦C<br>% | Weather station | To<br>Pr<br>Rho<br>WS | Temperature<br>Pressure<br>Humidity<br>Wind speed | ◦C<br>mm Hg<br>%<br>m/s |
| Office room | T4<br>RH4 | Temperature<br>Humidity | ◦C<br>% | | | | |
| Bathroom | T5<br>RH5 | Temperature<br>Humidity | ◦C<br>% | | Vis<br>TDE | Visibility<br>Tdevpoint | km<br>◦C |
| Outside | T6<br>RH6 | Temperature<br>Humidity | ◦C<br>% | Date | Ws<br>Day | Week status<br>Day of week | 1,0<br>[1,7] |
| Ironing room | T7<br>RH7 | Temperature<br>Humidity | oC<br>% | Whole system | AEC<br>LEC | Appliances energy consumption<br>Light energy consumption | Wh<br>Wh |

Table S2: Hyper-parameters of extreme gradient boosting (XGBoost) model

| # | Acronym | Description | Min | Max |
|---|---|---|---|---|
| 1 | $N_{est}$ | Number of estimators | 1 | 150 |
| 2 | $Max_d$ | Maximum depth of trees | 6 | 150 |
| 3 | $B$ | Type of booster | 0 (*gbtree*) | 1(*dart*) |
| 4 | $\eta$ | learning rate | 0 | 1 |
| 5 | $\gamma$ | Minimum loss reduction needed to create an extra partition on a leaf node | 0 | 1 |
| 6 | $Min_{cw}$ | Minimum sum of sample weight required in a leaf node | 1 | 10 |
| 7 | $sub_s$ | Sub-sample rate of the training set | 0 | 1 |
| 8 | $\lambda$ | $L2$ regularisation term on weights | 0 | 1 |
| 9 | $\alpha$ | $L1$ regularisation term on weights | 0 | 1 |

Table S3: The technical settings of the Machine learning methods

| # | Acronym | Full name | Hyper-parameters |
|---|---|---|---|
| 1 | KNN | K Nearest Neighbors | K=Number of neighbours |
| 2 | LoR | Logistic Regression | solver='lbfgs', penalty='l2',tol=0.0001, C=1.0, maxiter=100 |
| 3 | LR | Linear Regression | pre-defined settings (scikit-learn) |
| 4 | DT | Decision Tree Regressor | criterion='squared error', splitter='best', max depth= $D$, min samples split=2, min samples leaf=1, min weight fraction leaf=0.0, |
| 5 | MLP | Multi-layer Perceptron | solver='adam', activation='relu', alpha=1e-4, hidden layer sizes=(200,20,), max iter=1000 |
| 6 | BR | Bayesian Regression | Default settings |
| 7 | ET | Extra Tree | 'n estimators': 150, 'max depth': None, 'min samples split': 2, 'min samples leaf': 1, 'max features': None, 'bootstrap': False |
| 8 | AdaB | AdaBoost | number estimators=50, learning rate=1.0, loss='linear', base estimator='deprecated' |
| 9 | XGB | XGBoost | asymmetric trees, meaning splitting condition for each node across the same depth can differ |
| 10 | CBT | CatBoost | learning rate=0.01, iterations=15000 |
| 11 | LGBM | Light GBM | $num - leaves = 2^{max_{depth}}$, 'metric': 'rmse', 'num iterations':1000, 'num leaves': 100, 'learning rate': 0.001, 'feature fraction': 0.9, 'max depth': 100 |
| 12 | SVM | Support Vector Machine | kernel='rbf', C=1.0, epsilon=0.1, gamma='scale', and shrinking=True |
| 13 | DNN | Dense Neural Network | configured in Keras with activation='relu', optimizer='adam', loss='mean_squared_error', epochs=100, and batch size=32 |
| 14 | CDNN | Convolutional DNN | Conv2D layers with filters=64, kernel size=(3,3), activation='relu', optimiser='adam', and trained using loss='mean squared error', epochs=100, and batch size=32 |
| 15 | HGBR | Histogram-Based Gradient Boosting Regressor | learning_rate=0.1, max iter=100, max depth=6, l2 regularization=0.0, max bins=255 |

Table S4: The technical settings of the Machine learning methods

| | Sub-learners | Meta-learner | Stacking Model |
|---|---|---|---|
| 1 | ExtraTree +LGBM+RF+KNN | CATB | ST-M1 |
| 2 | ExtraTree+LGBM+RF+KNN | Linear | ST-M2 |
| 3 | ExtraTree+LGBM+RF+KNN | MLP | ST-M3 |
| 4 | ExtraTree+LGBM+RF+KNN+XGB | CATB | ST-M4 |
| 5 | ExtraTree+LGBM+RF+KNN+XGB | KNN | ST-M5 |
| 6 | ExtraTree+LGBM+RF+KNN+XGB | Linear | ST-M6 |
| 7 | ExtraTree+LGBM+RF+KNN+XGB | RF | ST-M7 |
| 8 | ExtraTree+LGBM+RF+KNN+XGB | SVM | ST-M8 |
| 9 | ExtraTree+LGBM+RF+KNN+XGB | XGB | ST-M9 |
| 10 | ExtraTree+LGBM+RF+KNN+XGB | ExtraTree | ST-M10 |

Table S5: Statistical analysis results of the appliances power consumption prediction using four proposed neuro-evolutionary methods.

| | XGB-PSO | | | | | | | XGB-GA | | | | | |
|---|---|---|---|---|---|---|---|---|---|---|---|---|---|
| | RMSE | MAE | MSLE | SMAPE | EVS | R-value | | RMSE | MAE | MSLE | SMAPE | EVS | R-value |
| Min | 6.17E+01 | 2.87E+01 | 1.25E-01 | 2.32E+01 | 5.33E-01 | 7.30E-01 | Min | 6.31E+01 | 2.89E+01 | 1.30E-01 | 2.31E+01 | 5.17E-01 | 7.24E-01 |
| Max | 7.34E+01 | 3.33E+01 | 1.51E-01 | 2.51E+01 | 5.96E-01 | 7.72E-01 | Max | 7.11E+01 | 3.18E+01 | 1.46E-01 | 2.46E+01 | 5.90E-01 | 7.68E-01 |
| Mean | 6.80E+01 | 3.09E+01 | 1.39E-01 | 2.41E+01 | 5.65E-01 | 7.52E-01 | Mean | 6.77E+01 | 3.06E+01 | 1.37E-01 | 2.39E+01 | 5.62E-01 | 7.50E-01 |
| Median | 6.81E+01 | 3.09E+01 | 1.39E-01 | 2.40E+01 | 5.64E-01 | 7.52E-01 | Median | 6.82E+01 | 3.06E+01 | 1.37E-01 | 2.40E+01 | 5.65E-01 | 7.52E-01 |
| STD | 3.09E+00 | 1.23E+00 | 7.20E-03 | 5.33E-01 | 1.80E-02 | 1.20E-02 | STD | 2.02E+00 | 7.24E-01 | 4.75E-03 | 3.64E-01 | 2.11E-02 | 1.34E-02 |

|  | XGB-EA | | | | | | | XGB-DE | | | | | |
|---|---|---|---|---|---|---|---|---|---|---|---|---|---|
|  | RMSE | MAE | MSLE | SMAPE | EVS | R-value |  | RMSE | MAE | MSLE | SMAPE | EVS | R-value |
| Min | 6.17E+01 | 2.83E+01 | 1.28E-01 | 2.32E+01 | 5.39E-01 | 7.34E-01 | Min | 6.24E+01 | 2.93E+01 | 1.31E-01 | 2.34E+01 | 5.47E-01 | 7.40E-01 |
| Max | 7.12E+01 | 3.28E+01 | 1.50E-01 | 2.52E+01 | 6.04E-01 | 7.77E-01 | Max | 6.99E+01 | 3.20E+01 | 1.42E-01 | 2.44E+01 | 5.98E-01 | 7.74E-01 |
| Mean | 6.79E+01 | 3.10E+01 | 1.39E-01 | 2.41E+01 | 5.67E-01 | 7.53E-01 | Mean | 6.63E+01 | 3.04E+01 | 1.37E-01 | 2.40E+01 | 5.69E-01 | 7.55E-01 |
| Median | 6.78E+01 | 3.11E+01 | 1.40E-01 | 2.42E+01 | 5.66E-01 | 7.53E-01 | Median | 6.67E+01 | 3.03E+01 | 1.37E-01 | 2.40E+01 | 5.67E-01 | 7.53E-01 |
| STD | 2.64E+00 | 1.19E+00 | 6.44E-03 | 5.03E-01 | 1.63E-02 | 1.07E-02 | STD | 2.06E+00 | 7.27E-01 | 3.08E-03 | 2.75E-01 | 1.63E-02 | 1.07E-02 |

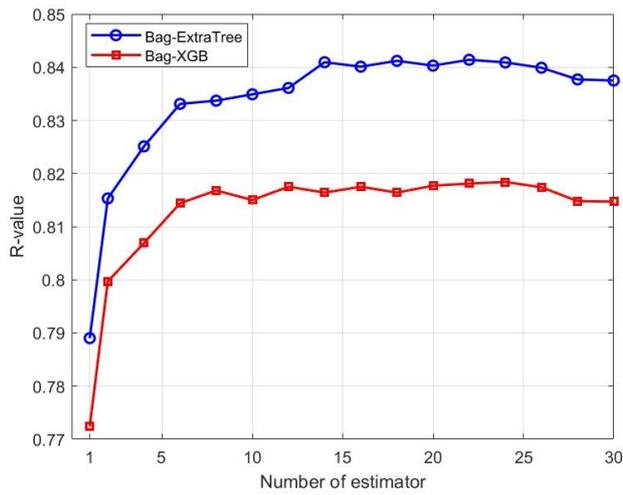 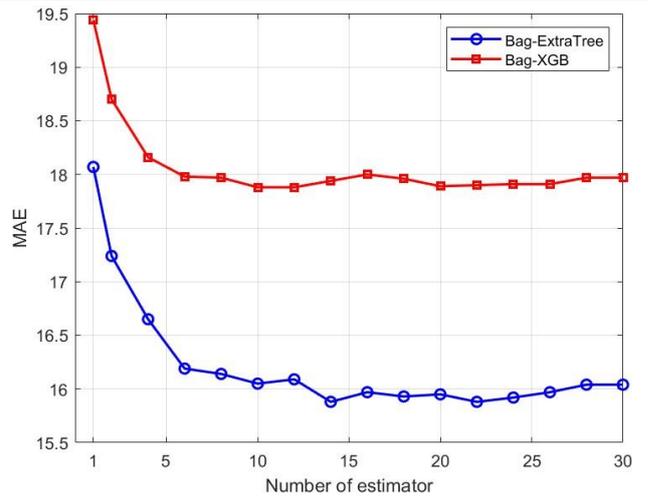

(a)                  (b)

Figure S1: Performance comparison of Bag-ExtraTree and Bag-XGB in terms of estimator number.

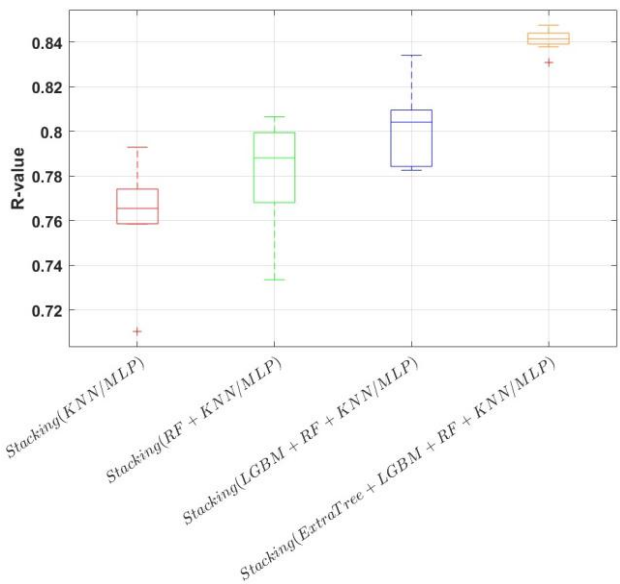 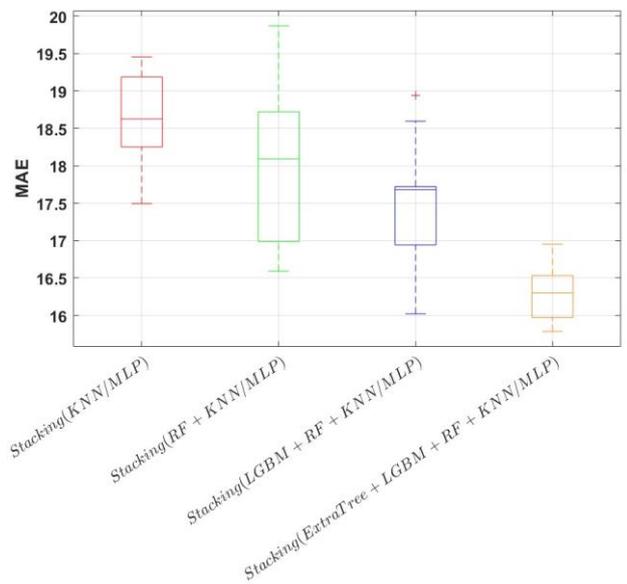

(a)                  (b)

Figure S2: Comparison of components impact on Stacking(ExtraTree+LGBM+RF+KNN/MLP) in terms of (a) accuracy and (b) MAE.